\documentclass[CGV,biber]{nowfnt} %

\usepackage[normalem]{ulem}

\usepackage{amsmath}
\usepackage{amssymb}
\usepackage{graphicx}
\usepackage{hyperref}
\usepackage{bm}
\usepackage{ulem}
\usepackage{mathtools}
\usepackage{tikz}

\usepackage{xcolor}
\definecolor{ibm1}{HTML}{648fff}
\definecolor{ibm2}{HTML}{785ef0}
\definecolor{ibm3}{HTML}{dc267f}
\definecolor{ibm4}{HTML}{fe6100}
\definecolor{ibm5}{HTML}{ffb000}
\definecolor{ibm6}{HTML}{000000}

\usetikzlibrary{positioning}

\newcommand{\EX}{\operatorname{\mathbb E}}

\DeclarePairedDelimiterX{\infdivx}[2]{[}{]}{%
  #1\;\|\;#2%
}
\DeclarePairedDelimiterX{\infdivy}[2]{[}{]}{%
  #1, #2%
}
\newcommand{\KL}{d_\text{KL}\infdivx}

\newcommand{\x}{\mathbf{x}}
\newcommand{\y}{{\mathbf y}}
\newcommand{\z}{\mathbf{z}}
\newcommand{\K}{\mathbf{K}}
\newcommand{\U}{\mathbf{U}}
\newcommand{\X}{\mathbf{X}}
\newcommand{\Y}{{\mathbf{Y}}}
\newcommand{\Z}{\mathbf{Z}}
\newcommand{\w}{\mathbf{w}}
\newcommand{\h}{\mathbf{h}}

\newcommand{\bu}{\mathbf{u}}

\newcommand{\bv}{\mathbf{v}}
\newcommand{\bk}{\mathbf{k}}
\newcommand{\bc}{\mathbf{c}}
\newcommand{\br}{\mathbf{r}}

\usepackage{mathtools}
\usepackage{cancel}
\usepackage[algoruled,algo2e]{algorithm2e}

\DeclarePairedDelimiter\ceil{\lceil}{\rceil}
\DeclarePairedDelimiter\floor{\lfloor}{\rfloor}
\DeclarePairedDelimiter\round{\lfloor}{\rceil}
\DeclarePairedDelimiter\Q{[\![}{]\!]} %

\newcommand\hatx{{\mathbf{\hat x}}}
\newcommand\haty{{\mathbf{\hat y}}}
\newcommand\hatz{{\mathbf{\hat z}}}

\newcommand\hath{{\mathbf{\hat h}}}
\newcommand\hatX{{\mathbf{\hat X}}}
\newcommand\hatY{{\mathbf{\hat Y}}}
\newcommand\hatZ{{\mathbf{\hat Z}}}

\newcommand\rate{{\mathcal R}}
\newcommand\pent{{P}} %
\newcommand\pdata{P_{data}} %

\newcommand{\setX}{\mathcal{X}}

\newcommand{\setW}{\mathcal{W}}

\newcommand{\setC}{\mathcal{C}}
\newcommand{\D}{\mathcal{D}}
\newcommand{\R}{{\mathcal R}}

\newcommand{\enc}{\texttt{e}}
\newcommand{\dec}{\texttt{d}}

\usepackage{xcolor}

\DeclareSourcemap{
\maps[datatype=bibtex]{
\map[overwrite]{
\step[fieldset=isbn, null]
\step[fieldset=issn, null]
\step[fieldsource=doi, final]
\step[fieldset=url, null]
}}}

\title{An Introduction to Neural Data Compression}

\maintitleauthorlist{
Yibo Yang \\
University of California, Irvine \\
yibo.yang@uci.edu
\and
Stephan Mandt \\
University of California, Irvine \\
mandt@uci.edu
\and
Lucas Theis \\
Google Research \\
theis@google.com
}

\issuesetup
{%
 copyrightowner={Y. Yang, S. Mandt and L. Theis},
 volume        = 15,
 issue         = 2,
 pubyear       = 2023,
 copyrightyear = 2023,
 isbn          = xxx-x-xxxxx-xxx-x,
 eisbn         = xxx-x-xxxxx-xxx-x,
 doi           = 10.1561/0600000107,
 firstpage     = 113, %
 lastpage      = 200
 }

\usepackage{mwe}

\author[1]{Yang,Yibo}
\author[2]{Mandt,Stephan}
\author[3]{Theis,Lucas}

\affil[1]{University of California, Irvine, USA; yibo.yang@uci.edu}
\affil[2]{University of California, Irvine, USA; mandt@uci.edu}
\affil[3]{Google Research, USA; theis@google.com}

\articledatabox{\nowfntstandardcitation}

\begin{document}

\makeabstracttitle

\begin{abstract}
Neural compression is the application of neural networks and other machine learning methods to data compression.
Recent advances in statistical machine learning have opened up new possibilities for data compression, allowing compression algorithms to be learned end-to-end from data using powerful generative models such as normalizing flows, variational autoencoders, diffusion probabilistic models, and generative adversarial networks. This monograph aims to introduce this field of research to a broader machine learning audience by reviewing the necessary background in information theory (e.g., entropy coding, rate-distortion theory) and computer vision (e.g., image quality assessment, perceptual metrics), and providing a curated guide through the essential ideas and methods in the literature thus far.
\end{abstract}

\chapter{Introduction}
\label{sec:intro} %

The goal of data compression is to reduce the number of bits needed to represent useful information. \textit{Neural}, or \textit{learned} compression, is the application of neural networks and related machine learning techniques to this task.
This monograph aims to serve as an entry point for machine learning researchers interested in compression by reviewing the prerequisite background and representative methods in neural compression.

\enlargethispage{-\baselineskip}
The basic idea of learning-based data compression has long existed in various forms before the current era of deep learning \cite{ziv1977universal}\cite{rissanen1979arithmetic}\cite{chou1989entropy}\cite{frey1997efficient}.
Many of the tools and techniques for neural compression, especially for images, also draw on a rich history of learning-based approaches in computer vision. 
Indeed, many problems in image processing and restoration can be viewed as lossy image compression; e.g., image super-resolution can be solved by learning a decoder for a fixed encoder (the image downsampling process) \cite{dong2015image}\cite{ledig2017srgan}.
In fact, neural networks have already been applied to image compression in the late 1980s and 1990s \cite{sonehara1989nn}\cite{sicuranza1990nn}, and even an early review article \cite{jiang1999image} exists. Compared to early work, modern methods differ markedly in their scale, neural architectures, and encoding schemes.

Current research in neural compression is heavily inspired by advances in deep generative modeling, such as GANs \cite{goodfellow2014gan}, VAEs \cite{kingma2014vae}\cite{rezende2015variational}, normalizing flows \cite{larochelle2011nade}, and autoregressive models \cite{theis2015lstms}\cite{vandenoord2016pixelrnn}.
While these models allow us to capture complex data distributions from samples (a key to neural compression), the research tends to focus on generating realistic data samples \cite{vandenoord2016wavenet}\cite{oord2017neural} or achieving high data log-density \cite{rezende2015variational}\cite{kingma2016improved}, objectives not necessarily aligned with data compression.

Arguably the first work exploring deep generative models for data compression appeared in 2016 \cite{gregor2016towards}, and the topic of neural compression has grown considerably since then. Multiple researchers have identified connections between variational inference and lossless \cite{frey1998bayesian}\cite{mackay2003information} as well as lossy \cite{balle2017end}\cite{theis2017cae}\cite{alemi2018fixing}\cite{yang2020improving} compression. 
This monograph hopes to further facilitate such exchange between these fields, raising awareness of compression as a fruitful application of generative modeling along with the associated challenges.

\begin{figure}[t]
    \centering
    \includegraphics[trim={0 4.2cm 0cm 0cm},clip,height=4cm]{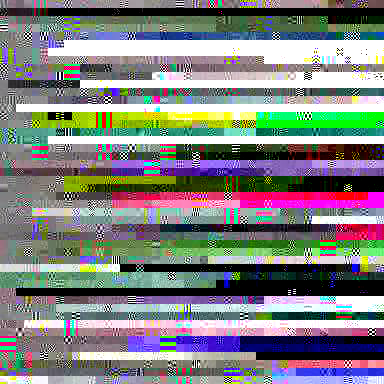}
    \includegraphics[trim={0 4.2cm 4.52cm 4.52cm},clip,height=4cm]{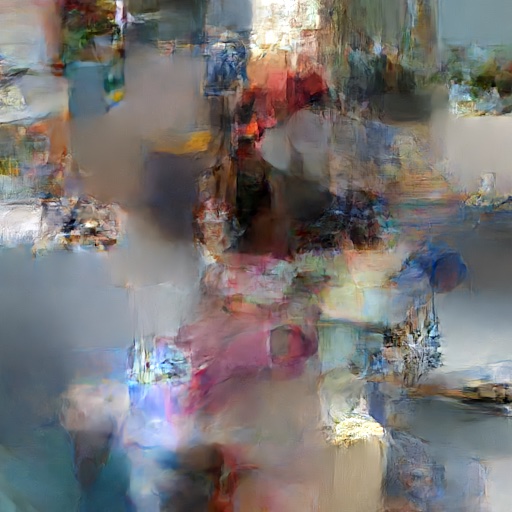}
    \caption{Compression as generative modeling. \textit{Left:} A sample drawn from the probabilistic model underlying JPEG, which betrays an assumption of independence among neighboring 8 by 8 pixel blocks (except for the DC components within each row). 
    \textit{Right:} A sample generated by a recent neural compression model by \citeauthor{minnen2020channel} \cite{minnen2020channel}.}
    \label{fig:jpeg}
\end{figure}

Instead of surveying the vast literature, we aim to cover the essential concepts and methods in neural compression, with a reader in mind who is versed in machine learning but not necessarily data compression. We hope to complement existing surveys that have a more specialized or applied focus \cite{balle2021ntc}\cite{ma2019image}\cite{liu2020deep} by highlighting the connections to generative modeling and machine learning in general.
In most of this monograph, we make essentially no assumption on the data other than that it is independently and identically distributed (i.i.d.), a typical setting for machine learning and statistics. We center our discussions around image compression, where most neural compression methods were first developed, but the basic ideas we present here are data agnostic. Towards the end, in Section~\ref{sec:video-compression}, we lift the i.i.d. assumption and consider video compression, which can be seen as an extension of the existing ideas along the temporal dimension. 

Neural compression can ease the development and optimization of data compression algorithms in a data-driven fashion. This can be especially useful for new or domain-specific data types, such as VR/AR content or scientific data, where developing custom codecs may otherwise be expensive.
Indeed, learning-based approaches are being applied to emerging data types, such as point clouds \cite{quach2019learning}\cite{guarda2019point}\cite{huang2020octsqueeze}, implicit 3D surfaces \cite{tang2020deep}, and neural radiance fields \cite{bird20213d}.
Effectively compressing such data may require new neural architectures \cite{tang2020deep} and/or domain knowledge to convert the data into neural-network-friendly representations \cite{huang2020octsqueeze}. 
However, the essential ideas and techniques introduced here for reducing the entropy, or bit-rate cost, of learned representations remain the same.

   \begin{figure}[t]
        \centering
        \input{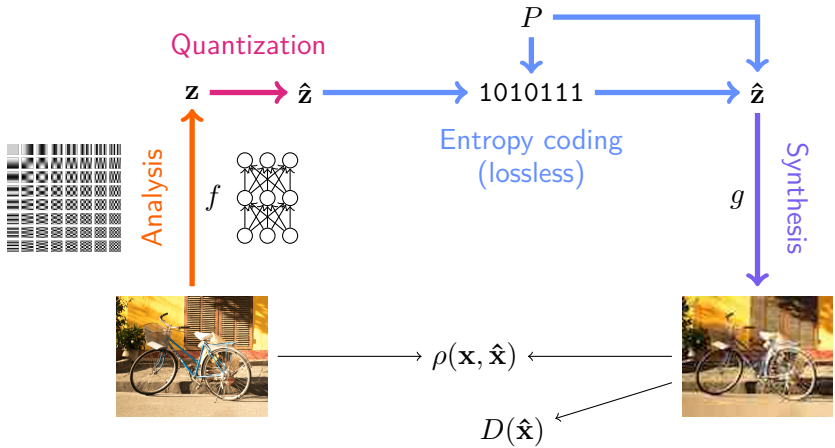}
        \caption{
        A typical pipeline in both neural and classical lossy image compression. An encoder transformation $f$ (for example, the DCT or a neural network) maps images to coefficients $\z$, which are first quantized to $\hatz$, and then entropy encoded into bits using an entropy model $P$. A reconstruction $\hatx$ is obtained using a decoder $g$ that aims for a small distortion $\rho$ between the data $\x$ and its lossy reconstruction $\hat{\x}$. In addition, neural compression can also involve an adversarial critic $D$, encouraging realism and high perceptual quality.
        }
        \label{fig:overview}
    \end{figure}

    \textbf{JPEG} \cite{itu1992jpeg} serves as a good motivating example of the lossy compression pipeline (depicted in Figure~\ref{fig:overview}). First introduced in 1992, it is still one of the most widely used image compression standards \cite{hudson2018jpeg}.
    At the heart of JPEG are linear mappings which losslessly transform pixels into coefficients and back. 
    The coefficients are first quantized to integers, incurring some information loss.
    Then they are further compressed losslessly by a combination of run-length encoding and entropy coding (the latter is discussed in Section~\ref{sec:entropy_coding}). 

\enlargethispage{-2\baselineskip}
    The linear portion of the encoding process consists of several steps. First, each pixel is  transformed from RGB to YCC coefficients 
    consisting of a luma component (Y) and two color components (C). After this color transform, each channel is treated independently, and optional downsampling is applied to the color channels. 
    Next, each channel is divided into $8 \times 8$ pixel blocks, and each block independently undergoes a \textit{discrete cosine transform} (DCT). 
    The transform coefficients are then linearly scaled and finally rounded to integers. 
    Given an image $\x$, the encoder thus performs
    \begin{align}
        \hatz = \lfloor \mathbf{D}\mathbf{A}\mathbf{C}\x\rceil,
    \end{align}
    where $\mathbf{C}$ is the pixelwise color transform, $\mathbf{A}$ is the block- and channelwise DCT, and $\mathbf{D}$ is a diagonal matrix scaling the coefficients. The decoder applies the transforms in reverse,
    \begin{align}
        \hatx = \mathbf{C}^{-1}\mathbf{A}^\top\mathbf{D}^{-1} \hatz.
    \end{align}
    Readers familiar with machine learning will be reminded of autoencoders \cite{bourlard1988auto}\cite{rumelhart1985learning} and it is natural to consider learned neural networks in place of the linear transforms. As we will see later, there are indeed close connections between lossy compression and variational autoencoders (VAEs) \cite{balle2017end}\cite{theis2017cae}\cite{alemi2018fixing}\cite{yang2022towards}, though other generative models have a role to play as well. What we call ``coefficients'' in the context of compression are often called  ``latent variables'' in the context of generative models. Like generative models, JPEG defines a probability distribution over coefficients which represents assumptions about the latent representation. Just as in VAEs, we can use this distribution to draw samples from the model underlying JPEG, with an example shown in Figure~\ref{fig:jpeg}.

    \textbf{Overview.} This introduction is organized into two main parts, lossless (Section~\ref{sec:lossless-compression}) and lossy (Section~\ref{sec:lossy-compression}) compression, with the latter relying on the former for compressing lossy representations of the data (see Figure~\ref{fig:overview}).
    We begin by reviewing basic \textit{coding theory} (Section \ref{sec:lossless-background}), and learn how we can turn the problem of lossless compression into learning a discrete data distribution, with the help of \textit{entropy-coding}.
    For this to work in practice, we decompose the potentially high-dimensional data distribution using tools from generative modeling, including \textit{auto\-regressive models} (Section \ref{sec:autoregressive}),  \textit{latent-variable models}, (Section \ref{sec:latent-variable-models}), and other models
    (Section \ref{sec:flow-and-other-models}).
    Each model class differs in its compatibility with different entropy-coding algorithms, and offers a different trade-off between the compression bit-rate and computational efficiency.
    \textit{Lossy} compression introduces additional desiderata, the most common being the \textit{distortion} of reconstructions, based on which the classical \textit{rate-distortion theory} and algorithms such as \textit{vector quantization} and \textit{transform coding} are reviewed (Section \ref{sec:lossy-background}).
    We then introduce \textit{neural lossy compression} as a natural extension of transform coding (Section \ref{sec:neural-lossy-compression}) and discuss the techniques necessary for end-to-end learning of quantized representations (Section \ref{sec:learning-quantized-representations-and-rate-control}), as well as lossy compression schemes that attempt to bypass quantization (Section \ref{sec:compression-without-quantization}). We then explore additional desiderata, such as the \textit{perceptual quality} of reconstructions (Section \ref{sec:perceptual-losses}), and the usefulness of learned representations for downstream tasks (Section \ref{sec:task-oriented-compression}), before briefly reviewing video compression (Section \ref{sec:video-compression}).
    Finally, we conclude in Section~\ref{sec:conclusion} with the challenges and open problems in neural compression that may drive its future advances.
    
    \chapter{Lossless Compression}
    \label{sec:lossless-compression}
    Lossless compression aims to represent data with as few bits as possible such that the data can be reconstructed perfectly.
    The basic recipe is to first build a probabilistic model of the data, and then feed its probabilities into a so-called entropy coding scheme which converts data into compact bit-strings. 
    This precludes non-likelihood-based models such as \textit{generative adversarial networks} (GANs) \cite{goodfellow2014gan}, from which it is hard to derive probabilities and which would often assign zero probability to data even if we could evaluate them.
    
    \section{Background}\label{sec:lossless-background}
    In the first few subsections, we review the basic concepts and algorithms for \textit{entropy coding}, which is the core interface between lossless compression and data modeling. Then in Section \ref{sec:dequantization}, we discuss a commonly used modeling technique for representing a discrete distribution with a density.

    \subsection{Entropy coding}
    \label{sec:entropy_coding}
    We call a sequence of outcomes of a discrete random variable a \textit{message}, and
    \textit{entropy coding} is a way to achieve lossless compression of a message. The basic idea, as embodied by \textit{symbol codes}, is to replace commonly occurring symbols with short codewords and rare symbols with longer codewords, thereby reducing the message's overall length. Morse code implements this idea by representing the common letter ``e'' with a single so-called \textit{dit} and the less frequent letter ``s'' using 3 \textit{dits}. To be able to distinguish between the message ``eee'' and the message ``s'', Morse code additionally requires pauses between encoded letters. These extra markers can be avoided by instead using a \textit{prefix-free} code where no codeword is the prefix of any other codeword \cite{cover2006elements}.

   By Shannon's source coding theorem \cite{Shannon1948}, the codeword length of an optimal prefix-free code is approximately the negative logarithm of the codeword's probability. Shannon also showed that the expected message length of an optimal prefix-free code is close to the \textit{entropy} of the message.

    \subsection{Entropy and information}
    The entropy\footnote{Entropy was originally defined as a thermodynamic concept by Clausius \cite{clausis1850entropy} and later Boltzmann \cite{boltzmann1895vorlesungen}, while the information theoretic entropy was introduced by Shannon~\cite{Shannon1948}.} of a discrete random variable $\X \sim P$ is a measure of uncertainty about its outcomes. It is based on the concept of \textit{surprise}, or \textit{information content}, which can be defined as the negative log-probability of an outcome $-\log_2 P(\x)$. Entropy, by definition, is the expected value of surprise, %
    \begin{align}
        H[\X] = \EX_{\x \sim P}[-\log_2 P(\x)]. \label{eq:entropy-definition}
    \end{align}
    For example, the entropy of a fair coin flip is 1 bit, and the entropy of a biased coin flip approaches 0 bits as the probability of one of the two outcomes approaches 1.
    
    When we losslessly compress data with a prefix-code, the entropy is the minimum number of bits required on average. More precisely,
    the expected message length $|C(\x)|$ of an optimal prefix-free code $C$ is bounded as
    \begin{align}
        H[\X] \leq \EX[|C(\X)|] \leq H[\X] + 1.
    \end{align}
    In practice, we typically do not know the data distribution $P$ but need to approximate it with a distribution $Q$. 
    We usually estimate $Q$ by maximum-likelihood, or equivalently, minimizing the \textit{cross-entropy} between $P$ and $Q$, defined as,
    \begin{align}
        H[P, Q] = \EX_{\x \sim P}[-\log_2 Q(\x)]. \label{eq:cross-entropy}
    \end{align}
    This is justified from a compression perspective, as the cross-entropy captures the average number of bits needed to code samples from $P$ using a code optimized for $Q$.
    The cross-entropy is always at least as large as the entropy, and the gap between the two is called \textit{relative entropy} or Kullback-Leibler (KL) divergence, defined as \begin{align}
        \label{eq:rkld}
        \KL{P}{Q} = \EX_{\x \sim P}[- \log_2 Q(\x) + \log_2 P(\x)].
    \end{align}
    It is always positive unless $P = Q$ and serves as an asymmetric distance measure between the two distributions.
    
    Based on these foundational concepts, we will review a few entropy coding schemes next.

    \begin{figure}
        \centering
        \begin{tikzpicture}[
    box/.style={anchor=south,rectangle,fill,gray,minimum width=0.8cm,inner sep=0,font=\color{white}\footnotesize\sffamily},
    huffnode/.style={draw,rounded corners,line width=1,fill=purple!10!white,rectangle,inner sep=1mm,minimum size=5mm,font=\color{purple}\footnotesize},
    edgelabel/.style={midway,fill=white,font=\footnotesize\sffamily,inner sep=0.5mm},
    edge/.style={line width=1},
    letter/.style={font=\small}
  ]
  \node[box,minimum height=0.90cm,label={below:\footnotesize\sffamily 0}] at (0cm, 0.7cm) {};
  \node[letter] at (0cm, 1.85cm) {`a'};
  \node[box,minimum height=0.34cm,label={below:\footnotesize\sffamily 100}] at (1cm, 0.7cm) {};
  \node[letter] at (1cm, 1.29cm) {`b'};
  \node[box,minimum height=0.32cm,label={below:\footnotesize\sffamily 101}] at (2cm, 0.7cm) {};
  \node[letter] at (2cm, 1.27cm) {`c'};
  \node[box,minimum height=0.24cm,label={below:\footnotesize\sffamily 110}] at (3cm, 0.7cm) {};
  \node[letter] at (3cm, 1.19cm) {`d'};
  \node[box,minimum height=0.20cm,label={below:\footnotesize\sffamily 111}] at (4cm, 0.7cm) {};
  \node[letter] at (4cm, 1.15cm) {`e'};
  
  \draw (-0.6cm, 0.7cm) -- (4.6cm, 0.7cm);

  \node[huffnode] (h1) at (0cm, 0cm) {$0.45$};
  \node[huffnode] (h2) at (1cm, 0cm) {$0.17$};
  \node[huffnode] (h3) at (2cm, 0cm) {$0.16$};
  \node[huffnode] (h4) at (3cm, 0cm) {$0.12$};
  \node[huffnode] (h5) at (4cm, 0cm) {$0.10$};

  \node[huffnode] (root) at (1.3cm, -2.8cm) {$1.0$};
  \node[huffnode] (b1) at (2.5cm, -1.9cm) {$0.55$};
  \node[huffnode] (b10) at (1.6cm, -1cm) {$0.33$};
  \node[huffnode] (b11) at (3.6cm, -1cm) {$0.22$};

  \draw[edge] (root) -- (h1) node[edgelabel] {0};
  \draw[edge] (root) -- (b1) node[edgelabel] {1};
  \draw[edge] (b1) -- (b10) node[edgelabel] {0};
  \draw[edge] (b1) -- (b11) node[edgelabel] {1};
  \draw[edge] (b10) -- (h2) node[edgelabel] {0};
  \draw[edge] (b10) -- (h3) node[edgelabel] {1};
  \draw[edge] (b11) -- (h4) node[edgelabel] {0};
  \draw[edge] (b11) -- (h5) node[edgelabel] {1};
\end{tikzpicture}
        \begin{tikzpicture}[
  box/.style={anchor=south,rectangle,draw=black,line width=0.5,minimum width=0.8cm,inner sep=0,font=\ttfamily\footnotesize},
  pbox/.style={anchor=south,rectangle,draw=black,line width=0.5,minimum width=0.8cm,inner sep=0,fill=purple!10!white,font=\color{purple}\footnotesize},
  letter/.style={font=\small},
  highlight/.style={fill=blue!10!white}
]

\node[pbox,minimum height=2.250cm] at (0cm, 0.000cm) {0.45};
\node[letter] at (-0.7cm, 1.125cm) {`a\textvisiblespace'};
\node[pbox,minimum height=0.850cm] at (0cm, 2.250cm) {0.17};
\node[letter] at (-0.7cm, 2.675cm) {`b\textvisiblespace'};
\node[pbox,minimum height=0.800cm] at (0cm, 3.100cm) {0.16};
\node[letter] at (-0.7cm, 3.500cm) {`c\textvisiblespace'};
\node[pbox,minimum height=0.600cm] at (0cm, 3.900cm) {0.12};
\node[letter] at (-0.7cm, 4.200cm) {`d\textvisiblespace'};
\node[pbox,minimum height=0.500cm] at (0cm, 4.500cm) {0.10};
\node[letter] at (-0.7cm, 4.750cm) {`e\textvisiblespace'};

\begin{scope}[xshift=0.8cm,yshift=0.0000cm]
\node[pbox,minimum height=1.0125cm] at (0cm, 0.0000cm) {};
\node[pbox,minimum height=0.3825cm] at (0cm, 1.0125cm) {};
\node[pbox,minimum height=0.3600cm] at (0cm, 1.3950cm) {};
\node[pbox,minimum height=0.2700cm] at (0cm, 1.7550cm) {};
\node[pbox,minimum height=0.2250cm] at (0cm, 2.0250cm) {};
\end{scope}

\begin{scope}[xshift=0.8cm,yshift=2.2500cm]
\node[pbox,minimum height=0.3825cm] at (0cm, 0.0000cm) {};
\node[pbox,minimum height=0.1445cm] at (0cm, 0.3825cm) {};
\node[pbox,minimum height=0.1360cm] at (0cm, 0.5270cm) {};
\node[pbox,minimum height=0.1020cm] at (0cm, 0.6630cm) {};
\node[pbox,minimum height=0.0850cm] at (0cm, 0.7650cm) {};
\end{scope}

\begin{scope}[xshift=0.8cm,yshift=3.1000cm]
\node[pbox,minimum height=0.3600cm] at (0cm, 0.0000cm) {};
\node[pbox,minimum height=0.1360cm] at (0cm, 0.3600cm) {};
\node[pbox,minimum height=0.1280cm] at (0cm, 0.4960cm) {};
\node[pbox,minimum height=0.0960cm] at (0cm, 0.6240cm) {};
\node[pbox,minimum height=0.0800cm] at (0cm, 0.7200cm) {};
\end{scope}

\begin{scope}[xshift=0.8cm,yshift=3.9000cm]
\node[pbox,minimum height=0.2700cm] at (0cm, 0.0000cm) {};
\node[pbox,minimum height=0.1020cm] at (0cm, 0.2700cm) {};
\node[pbox,minimum height=0.0960cm] at (0cm, 0.3720cm) {};
\node[pbox,minimum height=0.0720cm] at (0cm, 0.4680cm) {};
\node[pbox,minimum height=0.0600cm] at (0cm, 0.5400cm) {};
\end{scope}

\begin{scope}[xshift=0.8cm,yshift=4.5000cm]
\node[pbox,minimum height=0.2250cm] at (0cm, 0.0000cm) {};
\node[pbox,minimum height=0.0850cm] at (0cm, 0.2250cm) {};
\node[pbox,minimum height=0.0800cm] at (0cm, 0.3100cm) {};
\node[pbox,minimum height=0.0600cm] at (0cm, 0.3900cm) {};
\node[pbox,minimum height=0.0500cm] at (0cm, 0.4500cm) {};
\end{scope}

\begin{scope}[xshift=4.1cm]
\node[box,minimum height=2.500cm] at (0cm, 0.000cm) {0};
\node[highlight,box,minimum height=2.500cm] at (0cm, 2.500cm) {1};
\end{scope}

\begin{scope}[xshift=3.3cm,yshift=0.0000cm]
\node[box,minimum height=1.2500cm] at (0cm, 0.0000cm) {00};
\node[box,minimum height=1.2500cm] at (0cm, 1.2500cm) {01};
\end{scope}

\begin{scope}[xshift=3.3cm,yshift=2.5000cm]
\node[highlight,box,minimum height=1.2500cm] at (0cm, 0.0000cm) {10};
\node[box,minimum height=1.2500cm] at (0cm, 1.2500cm) {11};
\end{scope}

\begin{scope}[xshift=2.5cm,yshift=0.0000cm]
\node[box,minimum height=0.6250cm] at (0cm, 0.0000cm) {000};
\node[box,minimum height=0.6250cm] at (0cm, 0.6250cm) {001};
\end{scope}

\begin{scope}[xshift=2.5cm,yshift=1.2500cm]
\node[box,minimum height=0.6250cm] at (0cm, 0.0000cm) {010};
\node[box,minimum height=0.6250cm] at (0cm, 0.6250cm) {011};
\end{scope}

\begin{scope}[xshift=2.5cm,yshift=2.5000cm]
\node[box,minimum height=0.6250cm] at (0cm, 0.0000cm) {100};
\node[highlight,box,minimum height=0.6250cm] at (0cm, 0.6250cm) {101};
\end{scope}

\begin{scope}[xshift=2.5cm,yshift=3.7500cm]
\node[box,minimum height=0.6250cm] at (0cm, 0.0000cm) {110};
\node[box,minimum height=0.6250cm] at (0cm, 0.6250cm) {111};
\end{scope}

\begin{scope}[xshift=1.7cm,yshift=0.0000cm]
\node[box,minimum height=0.3125cm] at (0cm, 0.0000cm) {0000};
\node[box,minimum height=0.3125cm] at (0cm, 0.3125cm) {0001};
\end{scope}

\begin{scope}[xshift=1.7cm,yshift=0.6250cm]
\node[box,minimum height=0.3125cm] at (0cm, 0.0000cm) {0010};
\node[box,minimum height=0.3125cm] at (0cm, 0.3125cm) {0011};
\end{scope}

\begin{scope}[xshift=1.7cm,yshift=1.2500cm]
\node[box,minimum height=0.3125cm] at (0cm, 0.0000cm) {0100};
\node[box,minimum height=0.3125cm] at (0cm, 0.3125cm) {0101};
\end{scope}

\begin{scope}[xshift=1.7cm,yshift=1.8750cm]
\node[box,minimum height=0.3125cm] at (0cm, 0.0000cm) {0110};
\node[box,minimum height=0.3125cm] at (0cm, 0.3125cm) {0111};
\end{scope}

\begin{scope}[xshift=1.7cm,yshift=2.5000cm]
\node[box,minimum height=0.3125cm] at (0cm, 0.0000cm) {1000};
\node[box,minimum height=0.3125cm] at (0cm, 0.3125cm) {1001};
\end{scope}

\begin{scope}[xshift=1.7cm,yshift=3.1250cm]
\node[highlight,box,minimum height=0.3125cm] at (0cm, 0.0000cm) {1010};
\node[box,minimum height=0.3125cm] at (0cm, 0.3125cm) {1011};
\end{scope}

\begin{scope}[xshift=1.7cm,yshift=3.7500cm]
\node[box,minimum height=0.3125cm] at (0cm, 0.0000cm) {1100};
\node[box,minimum height=0.3125cm] at (0cm, 0.3125cm) {1101};
\end{scope}

\begin{scope}[xshift=1.7cm,yshift=4.3750cm]
\node[box,minimum height=0.3125cm] at (0cm, 0.0000cm) {1110};
\node[box,minimum height=0.3125cm] at (0cm, 0.3125cm) {1111};
\end{scope}
\end{tikzpicture}
        \caption{\textit{Left:} A Huffman tree over five symbols. Purple numbers in boxes denote probabilities, while binary strings above the leaf nodes correspond to the codewords assigned by the Huffman algorithm. \textit{Right:} Arithmetic coding of a message of length two. Each message corresponds to an interval whose length is proportional to the message's probability. As an example, any real number inside the interval corresponding to the message `ca' begins with \texttt{0.1010...} when written in binary form so that we can use \texttt{1010} to uniquely encode it. The message `aa' would be encoded as \texttt{000}.}
        \label{fig:huffman}
    \end{figure}
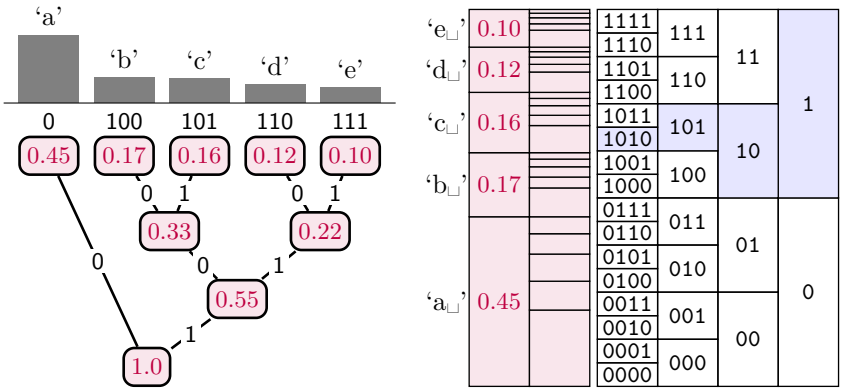

    \subsection{Huffman coding}
    Perhaps the  most well-known prefix-free symbol coding approach is Huffmann coding \cite{huffman1952method}.
    In a nutshell, given a distribution over symbols, the algorithm assigns unique binary codewords to every symbol by building a so-called Huffman tree (Figure~\ref{fig:huffman}). The leaf nodes of the tree are associated with the symbols. A Huffman tree can be recursively grown from the leaves to the root by successively merging the two nodes that have the lowest probabilities under the data distribution $P$. By traversing the tree from the root to a leaf, the sequence of branching directions (``0'' for left, ``1'' for right) assigns a unique binary codeword to each symbol, with the more frequent symbols receiving shorter codewords. Encoding and decoding are simple lookup operations with complexity linear in the tree depth.

    Huffman coding assigns a codeword with length at most $\lceil - \log_2$ $P(\x) \rceil$ for each symbol $\x$, and can be shown to be optimal among prefix codes \cite{cover2006elements}.
    However, as log-probabilities are generally not integers, Huffman coding (as with all so-called \textit{symbol codes}) incurs an overhead of up to 1 bit per symbol, relative to the information content. For example, a heavily biased coin with an entropy close to 0 will still require 1 bit to encode.
    When compressing a sequence of symbols, \textit{streaming codes} (see below) allow us to do so more efficiently than symbol codes by limiting the overhead to 1 bit for an entire \textit{message} rather than each symbol.

    \subsection{Arithmetic coding}
    \label{sec:arithmetic_coding}
\enlargethispage{\baselineskip}
    Arithmetic coding \cite{rissanen1979arithmetic}, also known as range coding \cite{martin1979range}, is our first example of a \textit{streaming code}. Streaming codes differ from symbol codes in that they assign codewords to entire messages and individual symbols do not have unique codewords. Streaming codes amortize the coding cost's overhead across the whole sequence of symbols and are therefore able to get closer to the entropy.

    The basic idea of arithmetic coding is to associate each symbol with a subinterval of the interval $[0,1]$ such that the subinterval's length equals the symbol's probability. This procedure can be generalized towards encoding symbols in sequence. For two symbols, the second symbol is no longer considered a subinterval of $[0,1]$ but a subinterval of the previous symbol's interval such that the second subinterval's length equals the two symbols' joint probability. For example, if symbol ``a'' has probability $1/5$ and is associated with $[0,0.2)$, the sequence ``aa'' would be encoded as $[0,0.04)$. This procedure can be iterated for sequences of arbitrary length, leading to a sequence of subintervals of decreasing size that contain each other. Any real-valued number contained in the final subinterval allows us to uniquely reconstruct the sequence of symbols and therefore encodes the entire sequence (Figure~\ref{fig:huffman}).
    
    In practice, one picks a representative number which can be represented with the smallest number of bits while uniquely identifying the interval. By construction, the size of the final sub-interval is the product of all previous symbols' conditional probabilities, hence the probability of the sequence. Intuitively, the interval's length (i.e., the sequence's joint probability) determines the number of relevant digits of the number representing the interval, thus the code length equals the log-probability of the sequence as required for entropy codes. If $\x^n$ is a message of length $n$, and $C(\x^n)$ the codeword assigned by arithmetic coding using an assumed probability distribution $Q$, then $|C(\x^n)| < \lceil -\log_2 Q(\x^n) \rceil + 1$ \cite{blelloch2013compression}. Thus if the true distribution of messages is $P$, then the average code length is at most $H[P, Q] + 2$.
    The advantage of arithmetic coding over Huffman coding is that the excess bits needed for encoding the message are amortized over all symbols, making it more efficient for long messages. Per symbol, the overhead is only $2/n$ bits, compared to $1$ bit per symbol for Huffman coding.
    
    Notably, arithmetic coding implements a first-in-first-out data structure, that is, a queue. Symbols which are encoded first are also decoded first. As we will see, this makes it a convenient choice for compression with autoregressive models (Section~\ref{sec:autoregressive}).

    \subsection{Asymmetric numeral systems}\label{sec:ans}
    More recent examples of streaming codes are asymmetric numeral systems (ANS) \cite{duda2009asymmetric}. While arithmetic coding implements a queue data structure, ANS operates like a stack. In arithmetic coding, the symbol encoded first is also decoded first (first-in-first-out) while ANS decodes the last symbol first (last-in-first-out).
    A detailed description is beyond our scope, and we follow Bamler \cite{bamler2022understanding} in our summary of the main idea (see Townsend \cite{townsend2020tutorial} for a different presentation).

    Numeral systems like the decimal or binary system can be interpreted as optimal codes for uniform distributions over a finite alphabet (the ``base'' of the numeral system). They encode a sequence of enumerated symbols into a single integer number, the ``stack''. To encode a symbol, we multiply the stack with the base (e.g., 2 or 10) and add the symbol. To decode, we recover the symbol as the stack modulo its base, while we shorten the stack by dividing it by the base. The stack's length in binary representation is approximately the number of symbols times the logarithm of the base of the numeral system, consistent with entropy coding. Interestingly, the stack's base can be changed from symbol to symbol and it is still possible to recover the sequence as long as the inverted order of bases is used upon decoding.

    ANS generalizes numeral systems from uniform probability distributions to non-uniform ones. Similar to arithmetic coding, symbols are first represented as subintervals of the unit interval. In addition, one discretizes the unit interval using a fine discretization grid. Every point on the grid belongs to the subinterval of a symbol and can be used to represent that symbol. The discretization points form a new alphabet of symbols that can be encoded using a numeral system, as described above. Since there are multiple discretization points in each subinterval, naively encoding one of them leads to a redundant code. However, ANS is able to avoid this redundancy. Bamler \cite{bamler2022understanding} notes that the mechanism by which ANS achieves this can be interpreted as a bits-back coding procedure, which will be discussed in Section \ref{sec:bits-back}. Like arithmetic coding, ANS incurs an overhead of up to 2 bits per message.

    \subsection{Continuous models for discrete data} \label{sec:dequantization}
    In this subsection, we show how a model for discrete data can be defined and parameterized by a continuous model, a common technique in generative modeling and neural compression.
    Lossless compression generally operates on discrete data. After all, it is infeasible to losslessly encode any real number with a finite number of bits. However, many generative models in machine learning assume continuous data, and model it with a density function. It turns out such continuous models are still useful for lossless compression.

    Suppose we have a density model $q$ over $\mathbb{R}^D$, and $\x \in \{0, \dots, 255\}^D$ is an RGB image following the ground truth image distribution $P$.
    We can derive a PMF over integers by integrating the density over hypercubes, as follows,
    \begin{align}
        Q(\x) := \int_{[-.5,.5)^D} q(\x + \mathbf{u}) \, d\mathbf{u}.
        \label{eq:discretized-density-model}
    \end{align}
    Assume we add uniform noise $\mathbf{u} \in [-0.5, 0.5)^D$ to the data and that the resulting noisy/dequantized data,  $\y = \x + \mathbf{u}$, has density $p$. 
    Then it is not difficult to show that \cite{theis2016note}
    \begin{align}
        -\int p(\y) \log_2 q(\y) \, d\y \geq -\sum_\x P(\x) \log_2 Q(\x).
        \label{eq:continuou-ll-rate-ub}
    \end{align}
    The left-hand side is the negative log-likelihood which we would optimize when fitting our generative model to the dequantized data. 
    Thus, we can minimize an upper bound on the lossless compression cost under the discretized model (RHS) by fitting a density via maximum likelihood (LHS), provided the discrete data is \textit{dequantized} appropriately with noise \cite{uria2013rnade}.
    
    The form of $Q$ as defined in Eq.~\ref{eq:discretized-density-model}, which we call a \textit{discretized density model}, is also useful in itself as a flexible model for discrete data. Examples include PixelCNN++ \cite{salimans2017pixelcnn++}, the prior distribution in a discrete flow \cite{hoogeboom2019integer}, as well as entropy models for neural compression \cite{mentzer2019practical}\cite{balle2018hyper}\cite{minnen2018joint}.
    
    The integral in Eq.~\ref{eq:discretized-density-model} in general is intractable to compute, but it can often be broken down into a series of univariate integrals.
    Therefore let us consider a univariate version of the discretized density model,
    
    \noindent
    \[
    Q_\theta(x) := \int_{[-0.5, 0.5)} q_\theta(x + u) d u.
    \]
    Let $F_\theta$ denote the CDF of $q_\theta$, then we can equivalently express the above in terms of a difference of CDF evaluations:
    \begin{align}
    Q_\theta(x) := F_\theta(x+0.5) - F_\theta(x-0.5) \label{eq:discretized-entropy-model-via-difference-of-CDFs}.
    \end{align}
    For example, if $q_\theta$ is the density of a logistic distribution, then $F_\theta$ is the logistic sigmoid function common in deep learning.

    \section{Autoregressive models}
    \label{sec:autoregressive}
    Autoregressive models exploit the fact that we can write any probability distribution as a product of conditional distributions using the chain rule of probabilities \cite{bishop2006PRML},
    \begin{align}
        \textstyle p(\x) = \prod_i p(x_i \mid \x_{< i}).
    \end{align}
    Here, $x_i$ is the $i$-th entry of the vector $\x$ and $\x_{< i}$ corresponds to all previous entries in an arbitrary order. The autoregressive factorization does not make any assumptions about the data distribution yet still allows us to easily incorporate useful assumptions. For example, a Markov assumption can be implemented by only considering the entries in $\x_{<i}$ which are close to $i$. A stationarity assumption is easily incorporated by using the same conditional distribution $p(x_i\,|\,\x_{< i})$ at every location. These two assumptions are often reasonable and can drastically reduce the amount of parameters of a model.

    Autoregressive modeling lends itself to lossless compression in combination with arithmetic coding (Section~\ref{sec:arithmetic_coding}) since both deal with data sequentially, one symbol at a time. Each symbol is encoded using the conditional distribution given the data that has already been encoded. This is in contrast to Markov random fields, for example, where conditional distributions are typically only tractable when conditioning on a larger neighborhood. Entropy coding generally requires the number of symbols in each encoding step to be manageable. Autoregressive models provide an important step towards practical entropy coding by decomposing a high-dimensional distribution into low-dimensional conditional distributions. Arithmetic coding is a better match for autoregressive models than ANS since it operates like a queue -- symbols encoded earlier will also be decoded earlier, and therefore will be available as input to conditional distributions.

    For data modalities such as audio or text, autoregressive models are an obvious choice due to the signal's sequential nature. Indeed, two of the top performing algorithms in the Large Text Compression Benchmark \cite{ltcb} use recurrent neural networks to predict the next token of a sentence. While \texttt{cmix} \cite{cmix} uses a mixture of long short-term memory networks (LSTMs) \cite{hochreiter1997lstm} and nonparametric models, \texttt{nncp} \cite{bellard2019nncp} relies solely on neural network models.
    However, autoregressive models have long also found application in image compression. 
    For example, JPEG \cite{itu1992jpeg} encodes the difference between neighboring DC coefficients, $z_{ij}^\text{DC} - z_{i(j - 1)}^\text{DC}$, in a raster-scan order, which can be thought of as implementing a first-order Markov model.

 \textit{Mixtures of experts} \cite{jacobs1991moe}\cite{hosseini2010cgsm}\cite{theis2012mcgsm} are a class of autoregressive models proven useful for compressing images,
    \begin{align}
        p(x_{ij} \mid \x_{<ij})
        = \sum_{k} \underbrace{p(k \mid \x_{<ij})}_\textsf{Gates} \underbrace{p(x_{ij} \mid \x_{<ij}, k)}_\textsf{Experts}.
    \end{align}
    
    The basic idea behind neural network extensions of this approach is to nonlinearly transform the inputs $\x_{<ij}$ before feeding them into the gates and the experts. For instance,
    RNADE \cite{uria2013rnade} used a fully connected neural network with a single rectified linear layer to transform the inputs $\x_{<ij}$. Other examples of deep autoregressive models include RIDE \cite{theis2015lstms}, PixelRNN \cite{vandenoord2016pixelcnn}, or PixelCNN++ \cite{salimans2017pixelcnn++}.
    More recent autoregressive modeling papers continue to explore different architectures for transforming $\x_{>ij}$. For example, the Image Transformer \cite{parmar2018it} uses an architecture based on self-attention \cite{vaswani2017attention}. PixelSNAIL \cite{chen2018pixelsnail} uses a combination of convolutions and self-attention layers. Glow \cite{kingma2018glow} uses invertible flows.

    With the exception of \texttt{nncp} \cite{bellard2019nncp} and \texttt{cmix} \cite{cmix}, all autoregressive models mentioned so far are \textit{static} models. That is, the models' parameters are trained once and then remain fixed during the entire encoding and decoding process. In contrast, a \textit{dynamic} model updates its parameters during the encoding process based on already encoded data. The decoder is then able to apply the same model updates based on the data already received. A simple dynamic autoregressive model for images was proposed by \citeauthor{wu1998p2ar} \cite{wu1998p2ar}. Here, the predictors are linear but the predictor's parameters are chosen dynamically. This is done by treating a larger neighborhood of preceding pixels as training data for the predictor. Meyer and Tischer \cite{meyer2001glicbawls} improved on this idea by weighting training points based on their distance to the pixel whose value we are predicting.

    Autoregressive models come with the restriction that decoding is an inherently sequential procedure. Each symbol can only be decoded after all the symbols have been decoded on which its prediction depends.
    To improve decoding speed, we can restrict the context $\x_{<i}$ to only a small neighborhood around $\x_i$, as for example in JPEG. Furthermore, the degree of parallelism can be increased by grouping coordinates of the data into blocks and only modeling the conditional dependencies between blocks, while treating data coordinates within each block as conditionally independent \cite{stern2018blockwise}\cite{minnen2020channel}.

    \section{Latent variable models}
    \label{sec:latent-variable-models}

    Latent variable models represent the data distribution using the sum rule of probability,
    \begin{align}
        p(\x) = \int p(\x, \z) ~ \text{d}\z = \int p(\x \mid \z) p(\z) ~ \text{d}\z,
        \label{eq:latent-variable-model}
    \end{align}
    where $\z$ is a vector of \textit{latent variables} (also called ``latents''), and the joint distribution $p(\x, \z)$ factorizes into a \textit{prior} $p(\z)$ and a \textit{likelihood} $p(\x\,|\,\z)$.
    Latent variable models are ubiquitous in machine learning and include hidden Markov models, mixture models \cite{bishop2006PRML}, and more recently variational autoencoders (VAEs) \cite{kingma2014vae}\cite{rezende2014vae}. 
    By integrating or summing over all possible realizations of the latent vector, latent variable models can capture complex dependencies in the data even when the prior and likelihood take simple forms.
    
    Training a latent variable model by (approximate) maximum-like\-li\-hood (see Eq.~\ref{eq:cross-entropy}) comes with a challenge: unlike in a fully-observed (e.g., autoregressive) model where the data probability $p(\x)$ can be readily evaluated, doing so is no longer straightforward since it is defined through an often intractable integral (Eq.~\ref{eq:latent-variable-model}). 
    Variational inference deals with exactly these complications, and we refer to \citeauthor{zhang2018advances} \cite{zhang2018advances} for more details on these methods. Here, we instead focus on the compression problem, which faces a similar challenge and uses similar tools. 
    Given a latent variable model consisting of a prior $p(\z)$ and likelihood $p(\x| \z)$, our goal is to compress a given data vector $\x$ with a code length close to its information content under the model, $-\log_2 p(\x)$.   To simplify the discussion below, we assume discrete $\z$ unless specified otherwise.

    \subsection{Two-part code}
    \label{sec:two-part-code}
    We start by considering a simple although generally suboptimal \textit{two-part code}~\cite{grunwald2007minimum}. Here, the data $\x$ is transmitted along with a latent vector $\z$ in two steps. First, the sender decides on a (ideally informative) vector $\z$, then encodes and transmits it under the prior $p(\z)$. 
    Next, $\x$ is compressed and transmitted under the likelihood model, $p(\x\,|\,\z)$. The receiver then decodes $\z$, and finally $\x$ given $\z$, using the same models.
    Both the prior and likelihood models are often fully factorized, allowing for coding each dimension of $\z$ or $\x$ in parallel, but can also be autoregressive models used in conjunction with arithmetic coding.
    Assuming negligible entropy coding overhead, the combined code length is
    \begin{align}
        l(\x, \z)  = -\log_2 p(\z) - \log_2 p(\x \mid \z) = -\log_2 p(\x, \z).
    \end{align}
    Moreover, the sender can (at least in principle) minimize this quantity over all choices of $\z$, resulting in the code length
    \begin{align}
    l_\text{MTP}(\x) = \min_{\z } \left( -\log_2 p(\z) - \log_2 p(\x \mid \z) \right). \label{eq:minimal-two-part-code}
    \end{align}
    The minimal two-part code length, $l_\text{MTP}(\x)$, is generally still suboptimal\footnote{However, it is possible to directly optimize a neural compression method for the two-part code. E.g., \citeauthor{mentzer2019practical} \cite{mentzer2019practical} trained a latent variable model with an inference network to minimize the expected coding cost of the two-part code, demonstrating much faster decoding speed than autoregressive models, although at $\sim 20\%$ worse bit-rate.} \cite{frey1998bayesian}\cite{wallace1990classification}\cite{honkela2004variational}. To see this, consider the following bound on the information content $-\log_2 p(\x)$,
    
\noindent
    \begin{align}
        -\log_2 p(\x) 
        &\leq -\log_2 p(\x) + \KL{q(\z \mid \x)}{p(\z \mid \x)} \\
        &= \mathbb{E}_{\z \sim q(\z \mid \x)}[ - \log_2 p(\z, \x)] - H[q(\z \mid \x)],
        \label{eq:nelbo}
    \end{align}
    also known as the \textit{negative evidence lower bound} (NELBO) of variational inference~\cite{zhang2018advances}.
    Here, $q$ is any distribution over $\z$ which may depend on $\x$. Crucially, the minimal two-part code length (Eq.~\ref{eq:minimal-two-part-code}) is a special case of the NELBO where 
    \begin{align}
        q(\z \mid \x) = \delta_{\z_\text{min}}(\z)
    \end{align}
    is a degenerate distribution centered on $\z_\text{min}$, which achieves the minimum in Eq.~\ref{eq:minimal-two-part-code}. In this case the entropy term vanishes. More generally, the NELBO is minimized when $q(\z \mid \x)$ equals the Bayesian \textit{posterior} distribution $p(\z \mid \x)$, the target of approximate inference \cite{zhang2018advances}.

    One may naturally wonder whether a coding cost equal to the NELBO can be realized for \textit{any choice of} $q(\z\,|\,\x)$.
    Bits-back coding, to be discussed next, answers this question in the affirmative. It further offers a compression interpretation of variational inference: for a given latent variable model, reducing the KL divergence between the distribution $q$ and the posterior distribution is equivalent to minimizing the code length of $\x$ under bits-back coding.

    \subsection{Bits-back coding}
    \label{sec:bits-back}
    While the optimal two-part code chooses a best latent vector $\z$ using deterministic optimization,
    bits-back coding \cite{frey1997efficient}\cite{wallace1990classification}\cite{hinton1993keeping}\cite{honkela2004variational} instead uses a \textit{stochastic} latent code in the form of a sample $\z$ from a distribution $q$. Surprisingly, this stochastic code can save bits compared to a deterministic code by allowing auxiliary information to piggyback on the choice of the latent code.
    To gain some intuition, consider again the optimization in Eq.~\ref{eq:minimal-two-part-code} for picking the optimal two-part code. When the given data $\x$ is uninformative of what latent variable $\z$ generated it, 
    the minimization over $\z$ may find multiple candidates that are nearly optimal. In the extreme case, ties need to be broken to settle on one of many equally good candidates.
    Instead of choosing a $\z_i$ randomly, the encoder can do so intentionally so as to communicate additional information via the chosen index $i$, provided that the decoder can reverse-engineer the encoding procedure. 

    Before proceeding, we emphasize a connection between sampling and decoding random bits. Given an entropy code derived from a discrete distribution, using it to \textit{decode} a sequence of uniformly random bits produces a random sample from this distribution.  For example, consider decoding a random bit-string according to a Huffman tree (Figure~\ref{fig:huffman}). As we traverse from the root to any leaf node, every edge along the path corresponds to a decision according to a coin flip. The probability of ending up with a particular symbol $\x$ with code length $|C(\x)|$ is $2^{-|C(\x)|}$, or approximately $2^{\log_2 p(\x)} = p(\x)$.
    
    In bits-back coding, the sender first generates a sample $\z \sim q(\z\,|\,\x)$ by \textit{decoding} a random bit-string $\xi$. Next, the sender encodes $\x$ under $p(\x\,|\,\z)$, then $\z$ under $p(\z)$, and then transmits the resulting bits. The receiver begins by decoding $\z$ and $\x$ from the received bits just as in the two-part code, but then proceeds to recover the exact bit-string the sender used to generate $\z$. This can be done by \textit{encoding} $\z$ under the distribution $q(\z\,|\,\x)$, which we assume the receiver has access to.
    Thus, $-\log_2 q(\z|\x)$ extra bits of information are recovered by the decoder (in addition to  $\x$) ``free of charge'', giving the scheme its name ``bits-back'' coding.
    Finally, and crucially, we note that the initial bits used to generate $\z$ do not have to be truly random. Indeed, to be useful, they should be supplied from a well-compressed bit stream of  \textit{auxiliary information} that also needs to be transmitted. For example, if our application requires us  to transmit an image along with a thumbnail version of it (e.g., for fast preview), the initial bits can come from the bit-string of the thumbnail image encoded with JPEG. We will examine this issue in more detail shortly.

   \begin{figure}[t]
        \centering
        \includegraphics[width=0.6\columnwidth]{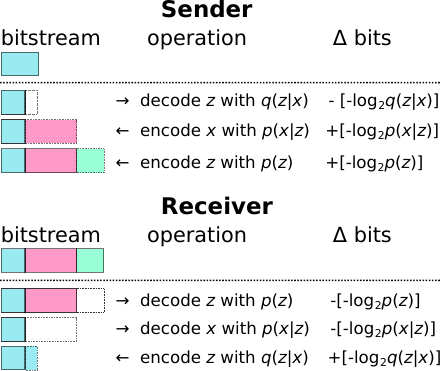}
        \caption{An illustration of the encoding (sender) and decoding (receiver) operations of bits-back coding. Auxiliary bits ($\xi$) are colored in blue.}
        \label{fig:bb-ans}
    \end{figure}

\enlargethispage{\baselineskip}  
    Figure~\ref{fig:bb-ans} graphically summarizes the algorithm.
    Given a bit-string $\xi$ of auxiliary bits and the data $\x$ to transmit, the sender starts by decoding a portion of $\xi$ into a stochastic latent code $\z$ of the data, ultimately sending a total number of bits equal to $|\xi| -\log_2 p(\z) - \log_2 p(\x|\z) + \log_2 q(\z|\x)$ to communicate both $\xi$ and $\x$. The net number of bits used for transmitting $\x$ alone is therefore
    \begin{align}
        -\log_2 p(\z) - \log_2 p(\x|\z) - (- \log_2 q(\z|\x))
    \end{align}
    which corresponds to a discount of $- \log_2 q(\z\,|\,\x)$ bits compared to a two-part code. Since $\z$ was drawn from $q(\z|\x)$, the expected coding cost is exactly the NELBO \cite{zhang2018advances}.

    The choice of auxiliary information is an important one in bits-back coding. 
    If the sender only wishes to communicate a single data sample $\x$ and no additional auxiliary information, the sender may set $\xi$ to be an artificial sequence of random bits, which can still be recovered by the receiver, but contains no useful information. In this case we do not get ``bits back'', and suffer an expected code length that is worse than the best two-part code.
    Even when there are auxiliary bits to transmit, the sender needs to commit to sending at least $- \log q(\z|\x)$ many of them
    for bits-back coding to achieve its nominal efficiency.
    This is known as the \textit{initial bits problem} \cite{frey1998bayesian}\cite{townsend2019practical}. 
    One potential solution is to compress a small portion of the data (e.g., a part of an image) with another codec, and use the resulting compressed bit-string for auxiliary bits \cite{townsend2019hilloc}.
    If multiple data samples need to be compressed, an elegant solution is  \textit{chaining} \cite{townsend2019practical} or ``bits-back with feedback'' \cite{frey1998bayesian}. The idea is to use the running bitstream of previously encoded data samples $\x_1, ..., \x_{i-1}$ as the source of auxiliary bits for encoding the sample $\x_i$. The initial bits problem becomes less severe with more data samples, and the code length per sample asymptotically converges to the NELBO.
    Chaining poses a requirement on the entropy coder that data is decoded in the exact opposite order in which it is encoded.
    Bits-back is therefore naturally combined with ANS coding which operates in stack order (Section~\ref{sec:entropy_coding}), popularized by the BB-ANS algorithm \cite{townsend2019practical}.

    \subsection{Improvements and extensions}
    \textbf{Continuous latents.}
    When the latent vector is continuous, bits-back coding can still be applied after discretizing the latent space and associated distributions \cite{mackay2003information}\cite{townsend2019practical}. More finely discretized latents will require more bits to encode, exacerbating the initial bits problem, but allow us to recover an equal amount of bits back. Bits-back coding thus allows us to code continuous latents with arbitrarily high precision.

    \textbf{Extension to models with multiple latents.}
    Although our discussion focused on a model with a single latent tensor, 
    \citeauthor{ho2019compression} \cite{ho2019compression} and \citeauthor{townsend2019hilloc} \cite{townsend2019hilloc} extended bits-back coding to hierarchical latent variable models, leveraging their superior expressiveness (in terms of better NELBO)
    to improve compression performance.
    Bit-Swap \cite{ho2019compression}, in particular, places restrictions on the latent hierarchy (e.g.,  Markovian) to allow recursive bits-back coding, alleviating the initial-bits problem. \citeauthor{ruan2021improving} \cite{ruan2021improving} and \citeauthor{townsend2021lossless} \cite{townsend2021lossless} developed bits-back schemes for sequential (e.g., time-series) latent variable models.

    \textbf{Iterative inference.}
    As discussed in Section \ref{sec:two-part-code}, the overhead of bits-back coding is equal to the KL divergence between $q(\z\,|\,\x)$ and the true posterior $p(\z\,|\,\x)$.
    The parameters of $q$ (e.g., mean of a Gaussian) are typically predicted from $\x$ by an \textit{amortized inference} network, as in a VAE (see, e.g., \cite{kingma2014vae}). Although cheap to compute, the resulting $q$ distribution is generally suboptimal \cite{cremer2018inference}. 
    \citeauthor{yang2020improving} \cite{yang2020improving} took a pre-trained model \cite{minnen2018joint} and proposed to directly minimize the KL gap with respect to the $q$ parameters for each given data point $\x$, resulting in improved image compression performance.

    \textbf{Extended latent spaces.}
    Analogous to the importance-weighted ELBO \cite{burda2015importance},  importance-weighted NELBO provides a tighter variational upper bound on the ideal code length than the NELBO (Eq.~\ref{eq:nelbo}). 
    \citeauthor{ruan2021improving} \cite{ruan2021improving} and \citeauthor{theis2021imortance} \cite{theis2021imortance} proposed bits-back coding schemes that operate in an extended latent space and operationalize the coding cost of the importance-weighted NELBO, and demonstrated improved compression performance.
    
    \section{Invertible flows and other models}
    \label{sec:flow-and-other-models}
    
    \textbf{Normalizing Flows.}
    Although \textit{continuous} normalizing flows do not lend themselves directly to lossless compression, local bits-back coding \cite{ho2019compression} allows lossless compression of finely quantized continuous data, with a code length close to the negative log data density up to a constant dependent on the quantization precision. The same method also allows losslessly compressing discrete data using a flow trained on the dequantization objective (e.g., the LHS of Eq.~\ref{eq:continuou-ll-rate-ub}). 
    By contrast, \textit{discrete normalizing flows} \cite{hoogeboom2019integer}\cite{tran2019discrete} directly model integer-valued data and support lossless compression. Such a flow learns a bijection $f$ to map data to an integer latent space, where a factorized prior is assumed; the given data $\x$ can then be compressed by simply entropy-coding its latent representation $f(\x)$, and decompressed using $f^{-1}$. This can be viewed as bits-back coding with a deterministic $q$. 

    \textbf{Other Models.}
    New types of generative models continue to be explored for lossless compression, aiming to achieve greater modeling power (hence better compression rate), while reducing computational demands. 
    For example, discrete diffusion models~\cite{sohl2015deep}\cite{hoogeboom2021autoregressive} offer parallel encoding/decoding and improved single-image compression rate. Furthermore, Probabilistic Circuits~\cite{peharz2020einsum} offer lightweight models and enable exact likelihood evaluation and efficient marginalization. \citeauthor{liu2021lossless} \cite{liu2021lossless} recently showed that this property makes them a compelling model for lossless compression. By proposing efficient en- and decoding algorithms, \citeauthor{liu2021lossless} achieved quasi-linear time complexity in the data dimension, resulting in an order of magnitude faster (de-)compression speed than available integer discrete flows or bits-back coding implementations. 

    \chapter{Lossy Compression}
    \label{sec:lossy-compression}
    In the following we review neural \textit{lossy} compression, that is, compression with imperfect data reconstruction. 
    When working with continuous data, such as an analogue signal, lossy compression is necessary, as it is impossible to losslessly store real values with a finite number of bits. 
    In other cases, a lossy reconstruction is often also sufficient, and the flexibility to discard ``irrelevant'' information allows us to achieve a far lower bit-rate than possible with lossless compression. 
    Lossy compression algorithms for digital media, such as audio, images, and video, routinely obtain an order of magnitude or more bit-rate savings than their lossless counterparts, without noticeable loss of quality to the end users. %

    We begin in Section~\ref{sec:lossy-background}  with basic background on lossy compression in the rate-distortion setting, reviewing the classical rate-distortion theory and algorithms such as vector quantization and transform coding. The rate-distortion VAE is introduced as a conceptual lossy compressor, embodying many approaches to be discussed. Section~\ref{sec:neural-lossy-compression} introduces neural lossy compression as a natural extension of transform coding, and discusses high-level architecture and modeling choices, with connections to hierarchical latent variable models. 
    We then discuss the core issue of end-to-end learning quantized representations and entropy modeling in Section~\ref{sec:learning-quantized-representations-and-rate-control}. 
    In Section~\ref{sec:compression-without-quantization}, we introduce an emerging technique of compression without quantization, which can be seen as operationalizing the bit-rate cost of a rate-distortion VAE, as given by a KL divergence.
    In aiming towards perceptually-pleasant reconstructions of data such as images, in Section~\ref{sec:perceptual-losses} we discuss various techniques for evaluating perceptual quality of images, and show how neural lossy compression can be optimized for high perceptual quality, in addition to the rate-distortion trade-off. We further consider the setting where the distortion is defined according to performance on downstream tasks in Section~\ref{sec:task-oriented-compression}, and finally give a brief overview of neural video compression in Section~\ref{sec:video-compression}.
    
    Frequently, we make reference to a \textit{quantization} operation, denoted by $\Q{\cdot}$. Quantization can be implemented in various ways in neural compression (e.g., by rounding to the nearest integer, denoted by $\round{\cdot}$; we give a detailed treatment in Section~\ref{sec:learning-quantized-representations-and-rate-control}), but at a high level is understood to be any mapping, either deterministic or stochastic, from a (often continuous) set of input values to a smaller and \textit{countable} set of output values. 

    \section{Background}
    \label{sec:lossy-background}

    We give a brief overview of the main results of lossy compression, in the classical rate-distortion setting studied by Shannon \cite{shannon1959coding}. Here, lossy compression involves encoding the given data into a \textit{discrete} representation, which is communicated losslessly to the receiver, using as few bits as possible.
    The quality of lossy reconstructions is measured by a fidelity criterion, or a \textit{distortion} function between pairs of original data samples and reconstructions. We refer readers to surveys by Berger \cite{berger1998lossy} and Gray and Heuhoff \cite{gray1998quantization} for more comprehensive coverage of classical results and historical developments in lossy compression.

    \subsection{Rate-distortion theory}\label{sec:rd-theory}
    For our purpose, we formalize a lossy compression algorithm (or a lossy \textit{codec}) as a 3-tuple consisting of an encoder, a decoder, and an entropy code, denoted by $c=(\enc, \dec, \gamma)$.
    The \textit{encoder} $\enc: \setX \to \setW$ maps each data point $\x \in \setX $ to a representation  $\w$ in a countable set $\setW$ (this can be the set of natural numbers, without loss of generality). %
    The \textit{decoder} $\dec: \setW \to \hat\setX$ maps each representation  $\w$ to a \textit{reconstruction point} $\hatx \in \hat\setX$.
    
    The encoder and decoder further agree to transmit the symbols of $\setW$ losslessly with an \textit{entropy code} $\gamma$ (see Section \ref{sec:entropy_coding}); we write $l(\x):=|\gamma(\enc(\x))|$ to denote the code length assigned to the encoding of $\x$.
    Suppose we are given a \textit{distortion function} $\rho: \setX \times \hat\setX \to [0, \infty)$ that measures the error caused by representing $\x$ with the lossy reconstruction $\hatx$. Historically, this is typically some form of a squared error, i.e., $\rho(\x, \hatx) \propto \|\x - \hatx\|^2$, and we will revisit this choice and discuss alternatives in Section \ref{sec:perceptual-losses}. Given the above, lossy compression is then generally concerned with  minimizing
    the average \textit{distortion}
    \begin{align}
        \D = \mathbb{E}_{\x \sim \pdata} [\rho(\x, \hatx)], \label{eq:rd-distortion} %
    \end{align}
    where $\hatx := \dec(\enc(\x))$,
    and simultaneously, the \textit{(bit-)rate}
    \begin{align}
        \R = \mathbb{E}_{\x \sim \pdata}[l(\x)], %
    \end{align}
    \looseness=-1 that is, the average number of bits needed to encode the data. We want to minimize these two quantities with respect to the choice of the encoding and decoding procedures $\enc$ and $\dec$, as well as the entropy code $\gamma$.

    Rate-distortion (R-D) theory \cite{shannon1959coding}\cite{cover2006elements} establishes limits on the performance of any such algorithm.
    Any lossy compression algorithm implements a noisy channel that receives a data input $\X$ and outputs a reconstruction $\hatX$, 
    described by a conditional distribution $Q_{\hatX | \X}$. The mutual information, $I[\X, \hatX]$, is then defined as the KL divergence between the joint distribution $\pdata \cdot Q_{\hatX | \X}$ and the product of its two marginal distributions. As we will also see in Section~\ref{sec:compression-without-quantization}, the mutual information $I[\X, \hatX]$ is a fundamental measure of the amount of information needed to transmit $\hatX$ given $\X$ (in bits per sample),
    and for a given distortion threshold $D$, the lowest achievable bit-rate is characterized by the information \textit{rate-distortion function} \cite{shannon1959coding},
    \begin{align}
        \R_I(D) = \inf_{Q_{\hatX | \X} : \mathbb{E}[\rho(\X, \hatX)] \leq D} I[\X, \hatX], \label{eq:info-rdf}.
    \end{align}
    The (information) R-D function is a fundamental quantity that depends only on the data distribution and choice of distortion function, and generalizes the Shannon entropy $H$ (Eq.~\ref{eq:entropy-definition}) from lossless compression to lossy compression.
    The R-D function has a few general properties, e.g., it is always non-increasing and convex, but is otherwise unknown analytically outside of a few cases. It can be numerically estimated by the Blahut-Arimoto algorithm \cite{blahut1972computation}\cite{arimoto1972algorithm} when the data and reconstructions are discrete and low-dimensional, and recent work has tackled the more general setting and estimated the R-D functions of continuous and high-dimensional data such as natural images \cite{yang2022towards}.

\begin{figure}
    \centering
    \includegraphics[width=0.55\columnwidth]{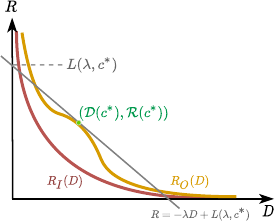}
    \caption{
    Visualizing the operational R-D optimization problem. We adjust our codec $c$ to minimize the $R$-axis intercept of a straight line (gray) with slope $-\lambda$ and passing through $(\D(c), \R(c))$; an optimum occurs when the line becomes tangent to the operational R-D curve $\R_O$ at the point $(\D(c^*), \R(c^*))$. Note that the operational R-D curve (orange) is not necessarily convex, and always lies above the information R-D curve $\R_I$ (red), i.e., $\R_O(D) \geq \R_I(D), \forall D.$
    }
    \label{fig:rd-optimization}
    \end{figure}

    In theory, the optimal rate $\R_I(D)$ is achievable by vector quantization \cite{cover2006elements}, to be described in Section~\ref{sec:vector-quantization}, but this approach to compression quickly becomes intractable for high-dimensional data. Rather than trying to achieve the theoretically optimal rate $\R_I(D)$ with any codec at any computational cost, in practice the design of a lossy codec is constrained by practical considerations, such as the computation budget of the target hardware, or the decoding latency acceptable for the application.   Denoting the set of all the acceptable codecs under consideration by $\setC$,  we can instead consider an \textit{operational} rate-distortion function\footnote{
    In usual treatments of R-D theory (see e.g., \cite{cover2006elements}), \emph{the} operational R-D function is defined as the asymptotic rate achievable by compressing multiple samples jointly using \textit{any} lossy codec, in the limit of infinitely many jointly compressed samples. This is different from our definition here, which instead focuses on the operational R-D performance achievable within a constrained family of codecs $\setC$, by compressing a single sample at a time (as is common in applications). Thus we generally only have $\R_O(D) \geq \R_I(D)$, i.e., $\R_I(D)$ is in general no longer achievable in this setup.
    }, formalized by
    \begin{align}
    \R_O(D) = \inf_{c \in \setC : \mathbb{E}[\rho(\X, \hatX)] \leq D} \mathbb{E}[l(\X)]. \label{eq:op-rdf}
    \end{align}
   Compared to Eq.~\ref{eq:info-rdf}, we replaced mutual information by the operational rate and optimize over an actual lossy codec $c$. 
   We can relax the constrained optimization problem to an unconstrained one, by introducing the rate-distortion Lagrangian \cite{blahut1972computation}\cite{chou1989entropy}, 
    \begin{multline}
    L(\lambda, c) = \mathcal{R}(c) + \lambda \mathcal{D}(c) = \mathbb{E}[l(\X)] + \lambda \mathbb{E}[\rho(\X, \hatX)]. \label{eq:operational-rd-lagrangian}
    \end{multline}
    For each fixed $\lambda > 0$, the minimum of this objective yields a 
    codec $c^*$ whose operational distortion-rate performance, $(\D(c^*), \R(c^*))$, lies on
    the convex hull of the operational R-D curve\footnote{In practice, due to non-convexity, we typically reach a local minimum of the Lagrangian, and therefore an R-D point lying above the operational R-D curve.}. This is illustrated in Figure~\ref{fig:rd-optimization}.
    The hyperparameter $\lambda$ can be interpreted as a Lagrange multiplier, and codecs with different operational rate-distortion trade-offs can be found by minimizing the Lagrangian with various $\lambda$. Most current end-to-end learned lossy compression methods to be discussed in Section~\ref{sec:neural-lossy-compression} follow this approach, training one codec for each $\lambda$. 
    We note that it is not always possible to attain every point on the operational R-D curve with this approach \cite{degrave2021tunable}, and alternative methods based on constrained optimization have been proposed  \cite{platt1988constrained}\cite{vanrozendaal2020lossy}. 
    
    \subsection{Connection to latent variable modeling and R-D VAEs}\label{sec:rd-vae}
    In Section \ref{sec:lossless-compression}, we saw that an integral part of lossless compression is estimating a good model of the data. In lossy compression, a similar connection can be drawn between rate-distortion optimization and data modeling, via a particular class of latent variable models  \cite{balle2017end}\cite{theis2017cae}\cite{alemi2018fixing}\cite{yang2022towards} which we will term  \textit{rate-distortion variational autoencoders}. %
    
    Let $\mathcal{Z}$ be any latent space, and $g: \mathcal{Z} \to \mathcal{\hat X}$ any measurable function. 
    Yang and Mandt \cite{yang2022towards} showed that an upper bound on the rate-distortion function $\R_I(D)$ can be obtained by optimizing the following objective:
    \begin{align}
        \mathcal{L}(q(\z | \x), p(\z), g, \lambda) := &\EX_{\x \sim \pdata}[ d_\text{KL}[q(\z|\x) \|p(\z)]] \nonumber \\
        & + \lambda \EX_{\x \sim \pdata,  \z \sim q(\z | \x)}[\rho(\x, g(\z))], \label{eq:rd-ub-lagrangian}
    \end{align}
    where $q(\z|\x)$ and $p(\z)$ are arbitrary distributions defined on $\mathcal{Z}$.
    The objective comes from a Lagrangian relaxation of the constrained optimization problem defining $\R_I(D)$ (in Eq.~\ref{eq:info-rdf}), with the expected KL divergence term serving as a variational upper bound on $I[\X, \hatX]$.
    Suppose there exists a conditional density \textit{aligned with} the given distortion function $\rho$, in the sense that 
    \[
    p(\x|\z) = C \exp\{ - \rho(\x, g(\z)\},
    \]
    where the normalizing constant $C$ does not depend on $\z$. In the common case of a squared error distortion, $p(\x|\z)$ is an isotropic Gaussian density with mean $g(\z)$ and constant variance.
    Then, the Lagrangian $\mathcal{L}$ can be easily seen to equal the NELBO objective of a ($\beta$-) VAE (up to a constant) \cite{balle2017end}\cite{theis2017cae}\cite{yang2022towards},
    \begin{align}
        \mathcal{L}(q(\z | \x), p(\z), g, \lambda) = &\EX_{\x \sim \pdata}[ d_\text{KL}[q(\z|\x) \|p(\z)]] \nonumber \\
        & + \lambda \EX_{\x \sim \pdata,  \z \sim q(\z | \x)}[- \log p(\x|\z)] + \text{const}, \label{eq:rd-vae-nelbo}
    \end{align}
    in which $p(\x|\z)$ is the likelihood/observation model, $q(\z |\x)$ is the variational posterior, $p(\z)$ is the (variational) prior, and $g$ is the decoder network.
    To make the correspondence with a VAE exact (``$\beta=1$''), we can also absorb $\lambda$ into the definition of the likelihood model, positing a density $p(\x|\z) = C \exp\{ - \lambda \rho(\x, g(\z)\}$. In the case of a squared distortion, this corresponds to a Gaussian density with a precision (inverse variance) parameter proportional to $\lambda$.
    Given the above equivalence between Eq.~\ref{eq:rd-ub-lagrangian} and Eq.~\ref{eq:rd-vae-nelbo}, we refer to the model associated with Eq.~\ref{eq:rd-ub-lagrangian} as a \textit{rate-distortion VAE} (or \textit{R-D VAE}),  even when there may not exist a proper likelihood density $p(\x|\z)$ aligned with the given distortion function $\rho$. 
    
    Besides yielding computationally tractable bounds on the R-D function \cite{yang2022towards}, rate-distortion VAEs also provide a useful perspective on many of the lossy compression algorithms we will encounter. Most existing neural lossy codecs can be seen as instances of R-D VAEs with a discrete latent space (Section~\ref{sec:ntc-connections-to-vaes-part1}), and many switch to a continuous latent space for end-to-end training \cite{balle2017end} (Section~\ref{sec:learning-quantized-representations-and-rate-control}), while methods that bypass quantization (Section~\ref{sec:compression-without-quantization}) almost always use a continuous latent space \cite{agustsson2020uq}\cite{theis2021algorithms}.

    \subsection{Vector Quantization}
    \label{sec:vector-quantization}
    \enlargethispage{-\baselineskip}
    Vector quantization (VQ), a classical technique from signal processing, is perhaps the most basic and general form of a lossy codec. 
    A vector quantization scheme, or a vector quantizer, consists of a set of integer representations $\setW = \{1, 2, ..., k\}$, an encoder $\enc$ assigning a given data to an integer representation $w \in \setW$, and a decoder $\dec$ that returns a reconstruction given $w$, usually implemented by indexing into a codebook of $k$ reconstruction vectors. 
    Note that there is no restriction on the encoding and decoding functions, other than the cardinality of $\setW$.
    The goal is then to determine an optimal quantization scheme for a given data distribution, under some objective. Commonly, the objective is to minimize a reconstruction error (Eq.~\ref{eq:rd-distortion}), as in the $k$-means algorithm, but can also more generally be an operational rate-distortion trade-off (Eq.~\ref{eq:op-rdf}).
    An optimal quantizer can be shown to always encode any given data point to its ``nearest-neighbor'' in the codebook to minimize the reconstruction error $\rho$ (or, a combination of reconstruction error and  code length, in the rate-constrained version), and set the $j$th entry in the codebook $\hatx_j$ to minimize the expected reconstruction error  between $\hatx_j$ and data points assigned to index $j$ \cite{chou1989entropy}.

    For some data source, e.g., the uniform or the Laplace distribution, optimal quantization can be characterized analytically \cite{ziv1985universal}\cite{sullivan1996efficient}.
    In most applications, Lloyd-Max-style algorithms (more widely known as $k$-means in machine learning) \cite{lloyd1982least}\cite{chou1989entropy} are instead used to estimate a quantization scheme from data samples.
    This usually involves minimizing an empirical rate-distortion cost (Eq.~\ref{eq:operational-rd-lagrangian}) over a dataset, which is still a basic ingredient in today's learned compression approaches \cite{balle2021ntc} (see Section \ref{sec:neural-lossy-compression}).

    Given unlimited data and compute, VQ can approximate any data distribution arbitrarily well. Indeed, the theoretical limit of lossy compression, the rate-distortion function $\R_I(D)$, can be shown to be achievable by jointly quantizing multiple data samples together with increasingly long blocks \cite{cover2006elements}.
    However, VQ comes with the severe downside of poor scalability and data efficiency \cite{gersho2012vector}.
    The computational and storage demand of VQ increases quickly with the data dimensionality.
    In high dimensions, an exceedingly large number of quantization points and high amounts of training data are needed to approximate the data distribution well. As a result, VQ is typically found in low-rate applications, such as low-rate speech coding \cite{sayood2012vectorq}. 
    We note however, $k$-means-style vector quantization has been successfully applied in a convolutional latent space for image generation \cite{oord2017neural}, and serves as an image tokenizer in many recent transformer-based models for image \cite{esser2021taming}\cite{yu2021vector} and text-to-image generation \cite{ramesh2021zero}\cite{yu2022scaling}.

    \subsection{Transform coding}
    \label{sec:transform-coding}
    Instead of quantizing the data directly, it is often much easier to do so in a transformed space, where the representation of the data is uncorrelated and scalar quantization can be effectively applied.
    The core idea behind transform coding \cite{goyal2001theoretical} is therefore to divide the task of lossy compression into decorrelation and quantization: first, the sender applies an \textit{analysis transform} $f$ to data $\x$, resulting in (ideally decorrelated) transform coefficients, $\z=f(\x)$; and second,
    coordinate-wise scalar quantization $\Q{\cdot}$ is applied to obtain a discretized representation $\hatz = \Q{\z}$.
    The symbols representing $\hatz$ can then be converted to a bit-string by entropy coding (discussed in Section~\ref{sec:entropy_coding}) under an entropy model $P(\hatz)$. The receiver losslessly recovers $\hatz$, and computes a reconstruction $\hatx = g(\hatz)$ using a \textit{synthesis transform} $g$, which is often the inverse of the analysis transform.
    In the terminology of Section~\ref{sec:rd-theory}, the encoder $\enc = \Q{\cdot} \circ f$ is the composition of the analysis transform $f$ followed by quantization, and the decoder $\dec$ is implemented by the synthesis transform $g$.

    The analysis transforms is typically linear and invertible, and is implemented as multiplication with an orthogonal matrix.
    A classic example is the Karhunen-Loève Transform (KLT), which is the same as principle component analysis (PCA) in our setting. The KLT maps data samples into the eigen-basis of the covariance matrix of the data distribution, and produces uncorrelated transform coefficients $\z$.
    Another example is the discrete cosine transform (DCT), which we saw in JPEG in Section \ref{sec:intro}. DCT can be shown to approximate PCA asymptotically \cite{Ahmed1974DiscreteCT}, and is one of the most widely used transforms in signal processing and compression.
    The choice of an orthogonal transform simplifies the theoretical analysis and design of transform coding algorithms, and for Gaussian data, the KLT followed by uniform scalar quantization can be shown to be optimal \cite{goyal2000transform}.

    Unlike vector quantization, which effectively optimizes over all possible choices of encoder $\enc: \setX \to \setW$ and decoder $\dec: \setW \to \hat\setX$, transform coding implicitly restricts the solution space of allowable codecs, e.g., the set of quantization points must be in the range of $g$, and therefore generally cannot achieve the unconstrained optimal performance of VQ. 
    The lack of theoretical optimality of transform coding is more than made up for by its vastly superior scalability  over VQ, as evidenced by its wide-spread use in the compression of digital media, such as images and videos \cite{goyal2001theoretical}. 
    \section{Neural lossy compression}
    
    \label{sec:neural-lossy-compression}
    Most current neural lossy compression methods are based on the paradigm of \textit{nonlinear transform coding} \cite{balle2021ntc} and use learned functions to encode data into a discrete representation, typically by  quantizing a continuous representation. 
    Compared to linear transforms (discussed in Section~\ref{sec:transform-coding}), non-linear transforms are more flexible and can better adapt to the data distribution \cite{balle2021ntc}, can be optimized for custom losses (such as perceptual losses in Section~\ref{sec:perceptual-losses}), and are demonstrated to be optimal in some cases \cite{wagner2021neural}. 
    In the following, we present high-level architectures and modeling choices, deferring a full discussion on model training and entropy modeling to Section~\ref{sec:learning-quantized-representations-and-rate-control}.

    \subsection{Overview}
\label{sec:3.2.1}
    In neural compression with non-linear transform coding, we replace the various components of transform coding (Section~\ref{sec:transform-coding}), such as the analysis transform $f$, synthesis transform $g$, and entropy model $P$, with neural networks or other function approximators, and learn them end-to-end on the data of interest.

    As in transform coding, we obtain a discrete representation $\hatz$ by quantizing the transform coefficients $\z$, i.e., $\hatz = \Q{\z}, \z = f(\x)$. It's common to refer to $\hatz$ as a tensor of ``latents'', due to connections to latent variable models (see Sections \ref{sec:rd-vae}, \ref{sec:ntc-connections-to-vaes-part1}, \ref{sec:ntc-connections-to-vaes-part2}), although there is some ambiguity in this terminology and some papers also apply it to the continuous transform coefficients $\z$. As before, an entropy model $\pent$ is used to losslessly transmit $\hatz$ via bit-strings (Section~\ref{sec:entropy_coding}). The entropy models are often based on powerful models from neural lossless compression (Section~\ref{sec:lossless-compression}).
    Most commonly, the objective is to simultaneously minimize the rate,
    \begin{align}
    \R := \EX[ - \log_2 \pent(\Q{f(\X)})],  \label{eq:ntc-rate-loss}
    \end{align}
    and the distortion,
    \[
    \D := \mathbb{E}[\rho(\X, g( \Q{f(\X)} )]. \label{eq:ntc-distortion-loss}
    \]
    The expectations are taken w.r.t. the data distribution $\pdata$ and are estimated with data samples.
    Unlike in Section~\ref{sec:rd-theory},
    here we no longer concern ourselves with an entropy code $\gamma$. Instead, we directly optimize the information content in place of a code length in Eq.~\ref{eq:ntc-rate-loss}, knowing that an entropy code can (in principle) always be derived from $P$ such that $-\log_2 P(\cdot) \approx |\gamma(\cdot)|$ (see Section~\ref{sec:entropy_coding}).
    If we denote the true marginal distribution of $\hatZ$ by $\pent^*$ (which is induced by $\pdata$ and the encoding procedure, via $\hatZ = \Q{f(\X)}$), then the rate loss can be equivalently written as the cross entropy,
    \begin{align}
    \rate = \EX_{\hatz \sim \pent^*(\hatz)}[ - \log_2 \pent(\hatz)],
    \label{eq:ntc-rate-loss-ce}
    \end{align}
    which precisely captures the cost of entropy coding $\hatz \sim \pent^*(\hatz)$ under our model $\pent(\hatz)$, and serves as an upper bound to the entropy $H[\hatZ]$.

    As in Eq.~\ref{eq:operational-rd-lagrangian}, we would like to optimize the operational R-D performance in the form of a rate-distortion Lagrangian, 
        \begin{align}
        \mathbb{E}[-\log_2 \pent( \Q{f(\X)} )]  + \lambda \mathbb{E}[\rho(\X, g( \Q{f(\X)} )], \label{eq:ntc-lagrangian}
    \end{align}
    for a suitable choice of the trade-off factor $\lambda$. However, as is written, the objective is not suitable for end-to-end training with SGD, due to the non-differentiability of the quantization operator and the rate loss.

    We dedicate Section \ref{sec:learning-quantized-representations-and-rate-control} to a detailed discussion of the various approaches to quantization, differentiable approximations, and the related topic of entropy modeling.
    In the rest of this section, we will examine other aspects and design choices of non-linear transform coding,

    \subsection{Connection to variational autoencoders}\label{sec:ntc-connections-to-vaes-part1}

    The R-D objective in Eq.~\ref{eq:ntc-lagrangian} closely resembles that of an autoencoder \cite{bourlard1988auto}\cite{rumelhart1985learning}, where the analysis and synthesis transforms $f$ and $g$ correspond to the encoder\footnote{Unfortunately, this use of the term ``encoder'' for the function $f$ of an autoencoder clashes with our definition of encoders $\enc$ in Section~\ref{sec:lossy-compression}. The ``encoder'' can also mean more abstractly the party initiating the data communication (and similarly, ``decoder'' can refer to the party receiving the data). The meaning is usually clear from the context.}  and decoder of an autoencoder. The rate term can be seen as a regularizer imposed on the representation $\hatz$, and differs from the regularization in traditional autoencoders based on dimensionality reduction or sparsity \cite{goodfellow2016deep}.
    
    Another way to interpret such a (deterministic) autoencoder is by viewing it as a \textit{variaitonal} autoencoder, but with a degenerate variational posterior distribution. Let us consider a rate-distortion VAE (introduced in Section~\ref{sec:rd-vae}) with discrete latent variables $\hatz$\footnote{The notation for latent variables in Section~\ref{sec:rd-vae} was ``$\z$'' instead. Unfortunately this collides with the use of ``$\z$'' as (unquantized) transform coefficients in this section.}, variational posterior $q(\hatz|\x)$ and prior $P(\hatz)$. For each fixed $\x$, if we define $q(\hatz|\x) := \delta_{\Q{f(\x)}}(\hatz)$, i.e., we let the approximate posterior concentrate all its mass on the quantized transform coefficients $\Q{f(\x)}$, then the R-D cost of non-linear transform coding in Eq.~\ref{eq:ntc-lagrangian} is equal to the NEBLO in Eq.~\ref{eq:rd-vae-nelbo}. Specifically, given a data sample $\x$, the bit-rate of the R-D VAE reduces to the code length in Eq.~\ref{eq:ntc-lagrangian}:
    
    \noindent
    \begin{align}
    d_\text{KL}[q(\hatz|\x) \| P(\hatz)] = - \log P(\Q{f(\x)}). \label{eq:discrete-kl-bit-rate}
    \end{align}
    This situation is analogous to the one in Section~\ref{sec:two-part-code}, where the two-part code can be considered a special case of bits-back coding with a deterministic posterior distribution. We achieve a coding cost equal to the KL-divergence (LHS of the above equation) by simply entropy coding the deterministic configuration of $q$ under model $P$.

    Do other kinds of R-D VAEs, especially ones with non-degenerate, or stochastic posterior distributions, also have a role to play in lossy compression algorithms? 
    We will address this question in Section~\ref{sec:ntc-connections-to-vaes-part2} and Section~\ref{sec:compression-without-quantization}.

    \subsection{Neural Transform Architectures}
    \textbf{Feedforward neural networks} are most often used for the encoding and decoding transforms $(f,g)$ in lossy compression, as in traditional autoencoders.
    For compressing unstructured data, fully-connected neural networks have been used \cite{balle2021ntc}\cite{yang2022towards}.
    In image compression, the networks are typically convolutional neural networks (CNNs), with $f$ implementing downsampled convolutions, and $g$ typically implementing upsampled convolutions  
    \cite{balle2016end} or sub-pixel convolutions \cite{theis2017cae}\cite{toderici2017full}.
    Various architectural improvements have been made to better capture and remove redundancies in the input data; these include replacing the usual ReLU activation with Generalized Divisive Normalization \cite{balle2016end}, introducing residual connections \cite{cheng2020learned}, attention mechanisms \cite{liu2019non}\cite{cheng2020learned}, and vision transformers \cite{zhu2021transformer}.
    
    We note that in the low-distortion / high-rate regime, a sufficiently large latent space \cite{yang2022towards} and transforms with sufficient capacity (e.g., as determined by the number of filters in a CNN) are generally needed to maximize the rate-distortion performance of a given architecture \cite{balle2017end}\cite{balle2021ntc}. Invertible networks have also been shown to provide good inductive bias for this setting \cite{helminger2020lossy}\cite{xie2021enhanced}. 

    \textbf{Recurrent and hierarchical architectures.} Instead of encoding data $\x$ into $\z$ with a single neural network, it can be beneficial to introduce a feedback mechanism to involve the decoder in the encoding process.
    We may divide the compression of $\x$ into $T$ stages, each stage generating an incremental representation $\hatz_t$ from the encoder to the decoder, and the final $\hatz$ consisting of the concatenation $\hatz_1, \hatz_2, ..., \hatz_T$.
    At the beginning of stage  $t$, the decoder has available a tentative data reconstruction $\hatx_{t-1}$ computed from stage $t-1$, using the already received $\hatz_1, \hatz_2, ..., \hatz_{t-1}$. Crucially, the encoder is equipped with a copy of the decoder, so having computed the same $\hatx_{t-1}$, the encoder only encodes the information in $\x$ that is not present in $\hatx_{t-1}$ (e.g., by encoding the \textit{residual}, $\x - \hatx_{t-1}$, in image compression). The resulting representation $\hatz_{t}$ is sent to the decoder, which then uses $\hatz_{t}$ to compute an improved tentative reconstruction $\hatx_t$. This process continues, with $\hatx_T$ declared the final data reconstruction $\hatx$. Here, $f:\x \to \hatz$ (similarly, $g$) is no longer implemented by a feedforward architecture, but rather a recurrent one, with $\hatx_t$ being the recurrent state.

    Such an approach embodies the idea of ``analysis-by-synthesis'', an influential model of perception and comprehension \cite{bever2010analysis}, and enables progressive compression whereby the data reconstruction $\hatx_t$ improves as more information in $\hatz_t$ is transmitted. 
    Many video compression methods in Section~\ref{sec:video-compression} also follow the same predictive coding paradigm, with $\x=[\x_1,...,\x_T]$ being a sequence of video frames, and $\hatz_t$ carrying the information in the reconstructed frame $\hatx_t$.

    We give a concrete example by \citeauthor{toderici2016rnn} \cite{toderici2016rnn}\cite{toderici2017full}, who proposed some of the first recurrent neural architectures for progressive and variable-rate image compression.
    Here, the computation at each stage can be summarized as
    \begin{align*}
        \hatz_t = \Q{ f_t(\br_{t -1}) }, \ & \br_t = \x - \hatx_t, \hatx_t = g_t(\hatz_t) + \alpha \hatx_{t-1} ; \\
        & \br_0 = \x, \hatx_0 = \mathbf{0} ,
    \end{align*}
    where $\br_t$, $f_t$, and $g_t$ denote the residual vector, encoding transform, and decoding transform at time $t$, respectively, and $\alpha \in \{0, 1\}$ allows two different modes of operation. With $\alpha=1$ (``additive reconstruction'' mode), $f_t$ and $g_t$ can simply be a pair of CNNs (with a separate pair for each $t$), and are trained to additively correct the cumulative reconstruction at each stage. With $\alpha=0$ (``one-shot reconstruction'' mode), $f_t$ and $g_t$ are chosen to be stateful, and typically LSTM architectures,  which are trained to directly predict the entire original image at each stage.
    Progressive and variable-rate image compression can then be achieved by controlling the number of steps $T$ of the recursive computation.

    The idea of reconstructing data progressively from abstract to increasingly specific concepts has also been well-established in deep generative modeling, often in the form of hierarchical latent variable models \cite{gregor2016towards} \cite{sonderby2016ladder} \cite{kingma2016improved}. Many of these generative models provide inspirations for compression. The recurrent architecture for compression discussed above is intimately related to bidirectional inference in hierarchical VAEs \cite{sonderby2016ladder}. Hierarchical and latent variable modeling is also prevalent in the entropy models of neural lossy codecs, in the form of a hyperprior \cite{balle2018hyper} and its extensions \cite{minnen2018joint, minnen2020channel}.
    Yang and Mandt \cite{yang2022towards} trained a deep ResNet-VAE \cite{kingma2016improved} to estimate bounds on the R-D function (see Section~\ref{sec:rd-vae}) and the theoretical performance of compression without quantization (to be discussed in Section~\ref{sec:compression-without-quantization}),
    while \citeauthor{duan2022lossy} \cite{duan2022lossy} trained a similar model with additive uniform noise (to be discussed in Section~\ref{sec:ntc-with-uniform-quantization}) to obtain improvements in practical lossy compression performance.

\enlargethispage{\baselineskip}
    \section{Learned quantization and rate control}\label{sec:learning-quantized-representations-and-rate-control}
    Neural networks have been used in image compression since before 1990 \cite{sonehara1989nn}\cite{jiang1999image}, but techniques were only recently developed to allow end-to-end training directly on a rate-distortion objective \cite{toderici2016rnn}\cite{balle2016end}\cite{theis2017cae}. The main stumbling block has been the fact that  the quantization operation and the discrete rate loss (see Eq.~\ref{eq:ntc-lagrangian}) are not differentiable.
    As quantization maps (usually continuous) input to a discrete set, its derivative is zero almost everywhere and undefined at points of discontinuity. 
    By the chain rule, the parameters of the encoder transform also receive zero gradient almost everywhere. Moreover,  
    the rate loss, defined via the PMF $P(\hatz)$, also has no derivative w.r.t. the discrete representation $\hatz$.
    
    Below we survey the major approaches developed over the years for dealing with the non-differentiability problem, organized by how quantization is done.  
    Since the choice of quantization affects the choice of the entropy model, as well as strategies for optimizing the bit-rate of quantized representations $\hatz$ (Eq.~\ref{eq:ntc-rate-loss-ce}), we discuss these issues jointly.
    
    We note that although most of these techniques are developed for lossy compression, they are equally useful in lossless compression methods that make use of quantization \cite{hoogeboom2019integer,mentzer2019practical}.

    \subsection{Binarization}
    Some of the earliest work in neural lossy compression obtained discrete representations $\hatz$ by binarizing the output of the encoder (analysis) network. The bit-rate was controlled in various ways when training the neural transforms; afterwards, a separate entropy model was fit to the empirical distribution of $\hatz$ and ultimately used for compression. We discuss specific approaches below.
    
    Based on techniques for training binarized neural networks \cite{williams1992simple}, 
    \citeauthor{toderici2016rnn} \cite{toderici2016rnn}\cite{toderici2017full} proposed stochastic binarization:  each scalar element $z$ of $\z$ is separately mapped to the interval $[-1, 1]$ (e.g., using point-wise $\operatorname{tanh}$ as the last layer of $f$), and then stochastically rounded to $-1$ or $1$ based on how close $z$ is to either value:
    \[
    \Q{z} := \operatorname{B}(z) = z + \epsilon, \quad \mathbb{P}(\epsilon) = \begin{cases} \frac{1+z}{2} \ , \ \epsilon = 1-z \\ \frac{1-z}{2} \ , \ \epsilon = -1-z \end{cases}
    \]
    To backpropagate through stochasitic binarization, they used a Straight-through Estimator (STE) \cite{bengio2013straight} and defined the gradient to be that of the identity function,
    \[
    \frac{d }{d z} \operatorname{B}(z) := \frac{d }{d z} \EX[\operatorname{B}(z)] = \frac{d }{d z} z = 1.
    \]
    \citeauthor{toderici2016rnn} \cite{toderici2016rnn}\cite{toderici2017full} trained the neural transforms $(f,g)$ to only minimize the distortion loss, relying on constraints on the dimensionality of the binary $\hatz$ to implicitly control the rate. After training, a separate autoregressive  entropy model (similar to a PixelRNN) is learned on the empirical distribution of $\hatz$ to further reduce the bit-rate \cite{toderici2017full}.

    \citeauthor{li2017learning} \cite{li2017learning} deterministically binarized $\z$ to $\{0, 1\}$ and used STE for backpropagation, optimizing a surrogate rate-distortion objective.  For rate control, they introduced a learned masking mechanism to the encoder network to encourage sparsity in $\hatz$, and optimized a surrogate rate loss defined in terms of the soft count of non-zeroed-out elements of $\hatz$. After training, they fit a separate PixelCNN-style entropy model to further reduce rate, similar to \citeauthor{toderici2017full} \cite{toderici2017full}.

    \subsection{Soft-to-Hard Vector Quantization}
    \citeauthor{agustsson2017soft} \cite{agustsson2017soft} proposed to use vector quantization with learned codebook values and introduced corresponding techniques for differentiable quantization and rate control. They considered the $k$-means-style (hard) quantization operation (see Section~\ref{sec:vector-quantization}), mapping $\z$ to its closest codebook vector,
    \[
    \Q{\z} := \operatorname{VQ}(\z, \setC) = \bc_j, \quad \text{ with } j = \arg \min_i \| \z - \bc_i \|,
    \]
    where $\setC = \{\bc_i | i=1,2,.., k\} $ is a finite set of codebook vectors in $\mathbb{R}^n$ learned alongside the model.
    \citeauthor{agustsson2017soft} \cite{agustsson2017soft} proposed to approximate hard quantization by a differentiable \textit{soft} quantization, via a linear combination of the codebook vectors weighted by how close they are to $\z$:
    \[
    \operatorname{VQ}(\z, \setC) \approx \operatorname{SoftQ}(\z, \setC) := \sum_i \phi_i \bc_i .
    \]
    Here $\phi \in \Delta^{N-1}$ is a probability vector computed as the softmax of weighted distances, $\phi = \phi(\z, \setC) := \operatorname{softmax}( -\sigma [\|\z - \bc_1\|^2, ...,  \|\z - \bc_M\|^2] )$, with $\sigma >0$ a hyperparameter.
    In practice, due to the prohibitive computation cost of VQ, the proposed method is only applied independently to small blocks of $\z$. 
    By annealing $\sigma \to \infty$ throughout training, soft quantization gradually approaches hard quantization, and a proper annealing schedule is needed to ensure effective training.
    For rate control, \citeauthor{agustsson2017soft} \cite{agustsson2017soft} optimized a surrogate rate loss based on the empirical distribution of the soft assignment probabilities, estimated via histograms.

    \citeauthor{mentzer2018conditional} \cite{mentzer2018conditional} simplified the above technique
    by performing only scalar quantization and dispensing with the annealing procedure. They fixed $\sigma$ at a constant (usually 1), and applied STE to differentiate through (hard) quantization using the gradient of soft quantization, \[
    \frac{\partial}{{\partial z}} \operatorname{VQ}(z, \setC):= \frac{\partial}{{\partial z}} \operatorname{SoftQ}(z, \setC).
    \]
    \citeauthor{mentzer2018conditional} \cite{mentzer2018conditional} furthermore trained a PixelCNN-style autoregressive entropy model end-to-end.
    To soften the non-differentiable discrete rate loss, they used the learned masking technique of \citeauthor{li2017learning} \cite{li2017learning}, but formed the surrogate rate loss based on the code length of non-zeroed-out elements of $\hatz$ under the concurrently trained autogressive entropy model, instead of naive counts as used by \citeauthor{li2017learning} \cite{li2017learning}.
    
    \subsection{Uniform Quantization (UQ)}\label{sec:ntc-with-uniform-quantization}
    
    Popularized by \citeauthor{balle2016end} \cite{balle2016end} and \citeauthor{theis2017cae}, \cite{theis2017cae}, 
    uniform quantization --- in its most common form --- rounds each element of $\z$ to the closest integer,
    \[
    \Q{\z} := \round{\z}.
    \]
    This can be viewed as a scalar version of
    the VQ approach, but with a fixed quantization grid equal to the set of integers.
    The assumption of a uniform  quantization grid with width 1  can generally be justified by using a sufficiently flexible pair of transforms $(f,g)$, which can
    warp the quantization grid in arbitrary ways if needed
    \cite{balle2016end}\cite{balle2021ntc}. 
    Compared to VQ, uniform quantization (and more generally, scalar quantization) is cheap to compute. Moreover, by embedding the integer-valued discrete representation $\hatz$ in $\mathbb{R}^{N}$, an entropy model can be conveniently specified in terms of a continuous density model, allowing for simpler differentiable rate surrogates than in approaches based on categorically distributed entropy models (e.g., \cite{agustsson2017soft}\cite{li2017learning}\cite{mentzer2018conditional}).
    Such an entropy model $P$ is defined by an underlying density $p$, exactly as in a discretized density model (Eq.~\ref{eq:discretized-density-model}),
    \begin{align}
        \pent(\hatz) := \int_{[-0.5, 0.5)^n} p(\hatz + \bv) d \bv,  \quad \forall \hatz \in \mathbb{Z}^n.  \label{eq:integer-entropy-model}
    \end{align}
    We defer details on the choice of $p$ to Section~\ref{sec:entropy-models}, and now discuss a few representative neural compression approaches based on this form of entropy model and integer-valued $\hatz$. In the rest of this sub-section, $\bu \sim \mathcal{U} \left([-0.5, 0.5)^n \right)$ is a random sample drawn from the uniform density $\mathcal{U}$ on the hypercube $[-0.5, 0.5)^n$.

    \textbf{UQ + STE.}
    \citeauthor{theis2017cae} \cite{theis2017cae}  proposed to train with uniform quantization and approximately differentiate through it by STE, using the identity gradient on the backward pass. For rate control, they optimized the same rate upper bound as on the LHS of Eq.~\ref{eq:continuou-ll-rate-ub}, replacing the code length $- \log_2 \pent(\hatz)$ by the differentiable upper bound $ \EX_\bu[ -\log_2 p(\hatz + \bu)]$.

    \textbf{Additive Uniform Noise.} \citeauthor{balle2016end} \cite{balle2016end} replace rounding with additive uniform noise for model training, i.e.,
    \[
    \round{\z} \approx \z + \bu, \; \bu \sim \mathcal{U} \left([-0.5, 0.5)^n \right).
    \]

    Naively, one might simply substitute the above into the Lagrangian Eq.~\ref{eq:ntc-lagrangian} and hope to obtain a reasonable training objective.
    However, the resulting code length, $-\log_2 \pent( f(\x) + \bu )$,  does not yet make sense, as our entropy model $\pent$ has only been defined over integers. It turns out the form of $\pent$ (Eq.~\ref{eq:integer-entropy-model}) offers a convenient solution:
    we can simply extend $\pent$ from $\mathbb{Z}^n$ to all of $\mathbb{R}^n$, by convolving the underlying density $p$ with the uniform noise $\bu$, i.e., 
    \begin{align}
    \tilde{p} := p * \mathcal{U} \left([-0.5, 0.5)^n \right). \label{eq:noisy-latent-density}
    \end{align}
    It's easy to see that $\tilde p$ agrees with $\pent$ on all integer points, and serves as a smoothed relaxation of $\pent$ which defines a surrogate gradient with respect to its input.
    Replacing $P$ by $\tilde p$ in Eq.~\ref{eq:ntc-lagrangian}, and taking expectation with respect to the uniform noise $\bu$, we obtain the surrogate training objective,
    \begin{align}
        \mathbb{E}_{\x \sim \pdata, \bu \sim \mathcal{U}}[-\log_2 \tilde p( f(\x) + \bu ) + \lambda \rho(\x, g( f(\x) + \bu ))], \label{eq:unoise-lagrangian}
    \end{align}
    which is now differentiable with respect to all components of the model, and can be simply estimated by Monte-Carlo sampling.

    \textbf{Stochastic Gumbel Annealing.}
    \citeauthor{yang2020improving} \cite{yang2020improving} proposed to optimize the following differentiable surrogate R-D objective:
    \begin{align}
        \EX_{\x \sim \pdata} \EX_{\hatz \sim q(\hatz|\x)}  [ -\log_2 \pent(\hatz) + \lambda  \rho (\x, g(\hatz)) ],
        \label{eq:disc-vae-rd-obj}
    \end{align}
    where $q(\hatz | \x)$ is an encoding distribution over integer valued latents, to be discussed below.
    Unlike the usual NELBO, the rate term above is a cross-entropy, rather than a KL divergence. The optimal choice of $q(\hatz | \x)$ is therefore deterministic, placing all of its mass on the choice of $\hatz$ that minimizes the R-D cost for each $\x$, in which case we recover a deterministic R-D VAE (Section \ref{sec:ntc-connections-to-vaes-part1}). Finding such an optimal deterministic encoder $q$ is a challenging discrete optimization problem, therefore the idea is to relax $q$ into a stochastic encoder (to enable gradient descent), and gradually anneal it towards a deterministic one.
    
    \citeauthor{yang2020improving} \cite{yang2020improving} parameterized $q(\hatz | \x)$ by a continuous location parameter $\z \in \mathbb{R}^n$ (e.g., predicted by an encoder network as $\z = f(\x)$) and a temperature hyperparameter $\tau > 0$. It is defined by
    \begin{align}
        q(\hatz | \x) &:= \prod_i q(\hatz_i | \x), \\
        q(\hatz_i | \x) & \propto \begin{cases}
            \exp\!\big\{-\psi\big(\z_i - \floor{\z_i}  \big)/\tau\big\} / C, & \quad\text{if $\hatz_i = \floor{\z_i}$}  \\[1pt]
            \exp\!\big\{-\psi\big(\ceil{\z_i} - \z_i \big)/\tau\big\} / C, & \quad\text{if $\hatz_i = \ceil{\z_i}$}
            \end{cases}
    \end{align}
    where $C$ is a normalizing constant, and $\psi = {\operatorname{tanh}^{-1}}$.
    The resulting categorical distribution concentrates its mass on the vertices of a hypercube containing $\z$, and generalizes stochastic binarization \cite{toderici2016rnn} to the integer lattice. 
    The probability mass assigned to each integer neighbor of $\z$ depends  inversely on its distance to $\z$, similar to the softmax formulation of \citeauthor{agustsson2017soft} \cite{agustsson2017soft}. The temperature hyperparameter is annealed towards zero over the course of optimization, such that sampling from $q$ converges to deterministic rounding $\z \to \round{\z}$. 
    For gradient-based optimization, the Gumbel-softmax trick \cite{jang2016categorical}\cite{maddison2016concrete} is used to differentiate through samples of $q$, and the discrete entropy model $P$ is replaced by its continuous extension $\tilde p$ (Eq.~\ref{eq:noisy-latent-density}), as in the uniform noise approach.
    
    This method, Stochastic Gumbel Annealing (SGA), was originally proposed to improve the compression performance of pre-trained models at test time \cite{yang2020improving}. In this work, $\z$ was initialized to the amortized prediction $f(\x)$, but then treated as a variational parameter, and iteratively optimized with gradient descent as in semi-amortized VI \cite{kim2018semi}.

    \citeauthor{tsubota2021comprehensive} \cite{tsubota2021comprehensive} further applied a version of SGA for end-to-end training, using STE instead of the Gumbel-softmax trick to differentiate through sampling $\hatz \sim q$ (which we refer to as SGA+STE), and obtained improved R-D performance compared to the UQ+STE approach.

    \textbf{Comparisons.}
    Empirical results \cite{Lee2019Context}\cite{minnen2020channel}\cite{agustsson2020scale} suggest that it is beneficial to train with different approximations for optimizing the distortion v.s. the rate terms of the R-D loss (Eq.~\ref{eq:ntc-lagrangian}). A recent empirical comparison of combinations of various approaches \cite{tsubota2021comprehensive} confirms this, showing that it is best to combine a rounding-based approximation (SGA+STE, UQ+STE) for the distortion term, and a uniform-noise-based approximation (additive uniform noise, dithered quantization \cite{choi2019uq}) for the rate term.

    \subsection{Connection to variational autoencoders, revisited} \label{sec:ntc-connections-to-vaes-part2}
    
    Previously in Section \ref{sec:ntc-connections-to-vaes-part1}, we interpreted a general non-linear transform coding model as a rate-distortion VAE (Section \ref{sec:rd-vae}) with a discrete latent space and deterministic variational posterior distribution. 
    In this section, we will consider other types of rate-distortion VAEs with continuous latent spaces, and discuss their relation to lossy compression.

    First, we show that the additive uniform noise approach for end-to-end training (discussed in Section~\ref{sec:ntc-with-uniform-quantization}) equivalently defines a rate-distortion VAE of the data, with a \textit{continuous} latent space and a particular choice of prior and posterior distributions. 
    To derive this, it is instructive to consider
    the density model $p$ as approximating the distribution 
    of the \textit{continuous} representation $\z = f(\x)$ as induced by $\x \sim \pdata$ and the analysis transform $f$. Indeed, suppose $\z$ is distributed according to $p$ (if our modeling is perfect), then the distribution of $\hat \z = \round{\z}$ has precisely the form of the entropy model $P$ defined by Eq.~\ref{eq:integer-entropy-model}. Moreover, if we define the ``noisy quantization'' by the random variable $\tilde \z := \z + \bu$, then its induced density (given $\z \sim p$) is precisely $\tilde{p}$ from Eq.~\ref{eq:noisy-latent-density}.
    Based on these connections, the surrogate Lagrangian from additive uniform noise (Eq.~\ref{eq:unoise-lagrangian}) defines a particular rate-distortion VAE, where $\tilde \z$ is the latent variable.
    Specifically, Eq.~\ref{eq:unoise-lagrangian} can be shown \cite{theis2017cae}\cite{balle2017end} to be equal to the NELBO objective of an R-D VAE, given by
     \begin{align}
         \EX_{\x \sim \pdata} \EX_{\tilde \z \sim q(\tilde \z|\x)} [ -\log_2 \tilde p(\tilde \z) + \log_2 q(\tilde \z|\x) - \log_2 p(\x|\tilde \z )] + \operatorname{const}, \label{eq:unoise-rdvae-nelbo} 
     \end{align}
    where $\tilde p(\tilde \z)$ plays the role of a prior, $q(\tilde \z | \x)$ is a fully factorized uniform posterior density centered at $\z= f(\x)$, and $p(\x|\tilde \z)$ is a likelihood density aligned with the distortion function $\rho$ \footnote{In the common case of a squared distortion, $p(\x|\tilde \z)$ is a Gaussian with mean equal to the reconstruction $\hatx = g(\tilde \z)$, and covariance inversely proportional to $\lambda$. See Section~\ref{sec:rd-vae}.}. 
    The equivalence can be seen by noting that the posterior entropy term is constant (in fact, 0), and sampling from the uniform posterior $q$ is equivalent to adding noise to $f(\x)$, by the reparameterization trick.

    Although additive uniform noise was originally motivated as a differentiable surrogate to the compression cost using deterministic uniform quantization (Eq.~\ref{eq:ntc-lagrangian}), the surrogate loss (Eq.~\ref{eq:unoise-lagrangian}) can be shown \cite{balle2021ntc}\cite{agustsson2020uq} to exactly equal the compression cost using \textit{universal quantization}, that is, quantization with a random offset, to be discussed in Section~\ref{sec:uq}.
    In other words, given our choice of posterior $q$ and prior $\tilde p$ distributions with particular shape restrictions, there is an efficient lossy compression procedure whose communication cost is fully differentiable (in particular, w.r.t. to the encoder parameters), with a bit-rate equal to the (expected) KL divergence $\EX_{\x \sim \pdata} [d_\text{KL}[q(\tilde \z|\x) \| \tilde p(\tilde \z)]]$.

    Next, we consider the possibility of using other types of R-D VAEs for lossy compression. 
    In generative modeling, the approximate posterior $q$ is typically chosen to be as flexible as possible to achieve a lower NELBO and learn a better model \cite{rezende2015variational}. It is therefore natural to wonder if other choices of the encoding distribution $q$ besides the uniform density can also be used to improve lossy compression.

    Given a data point $\x$ and an R-D VAE of the data, one idea is to simply quantize the mode or mean of the variational posterior $q(\z|\x)$, as in the uniform quantization approach of \citeauthor{balle2017end} \cite{balle2017end}, and entropy code the resulting quantized representation $\hatz$. 
    However, simple uniform quantization can give poor results, outside of the R-D VAE considered in Eq.~\ref{eq:unoise-rdvae-nelbo}. 
    \citeauthor{yang2020variational} \cite{yang2020variational} 
    proposed an improved quantization scheme for the case where $q$ is a factorized Gaussian with learned variances across different latent dimensions (as in a standard VAE \cite{kingma2014vae}). 
    In this method, the prior distribution $p(\z_i)$ is used to construct a quantization grid for each latent dimension $i$, taking inspirations from Arithmetic Coding. This is based on considering $k$-bit truncations of $\Z_i$ under the probability integral transform, for $k =1, 2, ...$, so the resulting set of grid points $\mathcal{G}_i$ consist of nested quantiles (median, quartiles, octiles, etc.) of $\Z_i$. 
    Then, given a data point $\x$ and its inferred posterior $q$, the posterior mean $\mu_i$ of each dimension $i$ is separately quantized to the corresponding grid $\mathcal{G}_i$, using a squared error distortion weighted inversely by the posterior variance $\sigma_i$ (this is essentially $-\log q(\z_i|\x_i)$ for a Gaussian $q$). The quantized value is found by efficiently solving a discrete optimization problem,
    \[
    \min_{\pi \in \mathcal{G}_i} \frac{ (\pi - \mu_i)^2 }{\sigma_i^2} + \lambda \rate(\pi),
    \]
    where $\rate(\pi)$ is the bit-rate associated with grid point $\pi$, and $\lambda$ is a rate-distortion trade-off hyperparameter shared across all latent dimensions. 
    The method therefore assigns more bits to latent dimensions with higher posterior uncertainty/variance, and fewer bits to latent dimensions where $d_\text{KL}[q(\z_i|\x)\|p(\z_i)]$ is small.
    Furthermore, variable bit-rate compression can be achieved by adjusting $\lambda$ at compression time, and \citeauthor{yang2020variational} showed that with this approach a single Gaussian VAE \cite{kingma2014vae} can already outperform JPEG in image compression \cite{yang2020variational}. However, this approach is outperformed by the end-to-end optimized approach \cite{balle2017end} at lower bit-rates, and does not operationalize the theoretical rate and distortion losses of the Gaussian VAE.
    
    Another idea is to find ways to transmit a sample of $q$ using close to $\EX_{\x \sim \pdata} [d_\text{KL}[q( \z|\x) \|  p(\z)]]$ bits (or, close to $d_\text{KL}[q( \z|\x) \|  p(\z)]$ bits for each given $\x$). This would allow us to operationalize the R-D loss of the associated R-D VAE, and more generally, yield a lossy compression scheme whose coding cost (as a KL divergence) can often be easily optimized with gradient descent. In theory, this approach can even attain the rate-distortion theoretic limit of lossy compression, up to a logarithmic overhead \cite{yang2022towards}, and can also be more efficient in terms of the rate-distortion-perception trade-off \cite{theis2021stochastic, theis2022diff}. However, an efficient implementation of this approach is non-trivial, and likely even impossible in the worst case \cite{agustsson2020uq}. We will discuss this approach, and the related topics, in Section~\ref{sec:compression-without-quantization}.

    \subsection{Entropy models}\label{sec:entropy-models}
    Various entropy models have been proposed to reduce the bit-rate and improve the rate-distortion performance of neural lossy compression, using largely the same modeling ideas as for lossless compression.
    
    Following quantization, various models can be used to further losslessly compress the resulting discrete representations --- for instance, autoregressive models  \cite{li2017learning}\cite{toderici2017full}\cite{mentzer2018conditional} or off-the-shelf adaptive entropy codes \cite{agustsson2017soft}. 
    The most common approach in end-to-end methods is to combine uniform quantization with an entropy model parameterized in terms of a density $p$ as in Eq.~\ref{eq:integer-entropy-model}. 
    The simplest choice is a fully-factorized $p$, resulting in a factorized entropy model. Each marginal of $p$ is typically parameterized as a mixture distribution \cite{theis2017cae}, or indirectly as the derivative of a deep CDF model \cite{balle2018hyper} (exploiting the relation in Eq.~\ref{eq:discretized-entropy-model-via-difference-of-CDFs}).
    
    Going beyond factorized entropy models, recent research has explored latent-variable modeling, autoregressive modeling, and their combination, to increase the flexibility of the prior density $p$ and improve the compression bit-rate \cite{balle2018hyper}\cite{minnen2018joint}\cite{minnen2020channel}.
    The basic latent variable model approach, commonly referred to as the \textit{hyperprior} approach \cite{balle2018hyper}, expresses the entropy model's underlying density through an additional hierarchy of latent variables $\h$ (``hyper-latents''),
    \begin{align*}
    \h &\sim p(\h), & \hath &= \round{\h}, \\
    \z \mid \hath &\sim p(\z \mid \hath), & \hatz &= \round{\z}.
    \end{align*}
    The hyperprior density, $p(\h)$, is typically parameterized as in a factorized entropy model, while $p(\z\,|\,\hath)$ is a density (e.g.,  factorized Gaussian) whose parameters are predicted from $\hath$ by a neural network (``hyper-decoder'').
    Crucially, note that the prior density of $\z$ is conditioned on the discrete $\hath$, as the hyper-latents must be discretized and entropy-coded first at compression time. 
    The information transmitted in the hyper-latents is known as \textit{side information} \cite{balle2018hyper}, and lets the sender and receiver dynamically select an entropy model based on the content of the input data.
    To train such an entropy model, the rate loss (Eq.~\ref{eq:ntc-rate-loss-ce}) is modified to account for the side-information, replacing $- \log_2 \pent(\hatz)$ by the joint information content $- \log_2 \pent(\hatz, \hath) = - \log_2 \pent(\hath) - \log_2 \pent(\hatz \,|\, \hath)$; the same techniques from Section~\ref{sec:ntc-with-uniform-quantization} can then be used to differentiate through quantization and rate loss.  
    Empirically, the hyperprior considerably reduces the overall bit-rate of a factorized entropy model, with the side-information comprising a small percentage of the overall rate \cite{balle2018hyper}. 

    The bit-rate can be further improved by additionally modeling $\hatz$ autoregressively similarly to a PixelCNN~\cite{minnen2018joint}, but results in serial and hence slower decoding. To address this, \citeauthor{minnen2020channel} \cite{minnen2020channel} proposed to instead use channelwise (instead of spatial) autoregressive conditioning, significantly speeding up (de)compression without harming the rate-distortion performance. Compared to an autoregressive model, a latent-variable entropy model has the advantage of parallel encoding/decoding via (conditionally) factorized distributions, but entails transmitting side-information, similar to the two-part code in lossless compression (Section~\ref{sec:two-part-code}). \citeauthor{yang2020improving} \cite{yang2020improving} applied bits-back coding to reduce the transmission of side-information in a hyperprior model. 

    Regardless of the choice of an entropy model, the sender and receiver must agree on the exact same probabilities for entropy coding (such as Arithmetic Coding, discussed in Section~\ref{sec:arithmetic_coding}) to operate correctly. This can be a stringent requirement when the entropy models (e.g., the conditional model $P(\hatz|\hath)$) are computed on the fly, especially in the face of round-off errors from floating point arithmetic and non-deterministic GPU operations. We refer readers to \citeauthor{balle2018integer} \cite{balle2018integer} for more details on this issue, and a potential solution based on integer arithmetic.
    
    \section{Compression without quantization}
    \label{sec:compression-without-quantization}
    The non-differentiability of quantization has hindered end-to-end training of lossy compression models. The various methods in Section~\ref{sec:learning-quantized-representations-and-rate-control} replace quantization by a differentiable surrogate at training time, leading to a mismatch between ``soft'' quantization during training and ``hard'' quantization at test time. This mismatch generally results in sub-optimal performance \cite{yang2020improving}.
    Annealing can alleviate the problem \cite{agustsson2017soft}\cite{yang2020improving}\cite{agustsson2020uq} but may suffer high-variance gradients as the approximation approaches hard quantization, and requires specifying an annealing schedule.

    Perhaps surprisingly, quantization can be avoided entirely if we are willing to accept some noise in the transmitted representation. Instead of quantization, it is possible to transmit a continuous but stochastic sample $\z \sim q(\z \mid \x)$ using a finite number of bits \cite{bennett2002reverse}. The problem of efficiently communicating such a sample is also recognized as \textit{channel simulation} \cite{zamir2014book}\cite{li2017strong}, \textit{reverse channel coding} \cite{bennett2002reverse}\cite{agustsson2020uq}\cite{theis2021algorithms}, or \textit{relative entropy coding} \cite{flamich2020cwq}\cite{flamich2022astar}.

\enlargethispage{\baselineskip}
    Bits-back coding (Section~\ref{sec:bits-back}) seems like a natural candidate for this problem as it also uses a stochastic encoder. Unfortunately, a requirement of bits-back coding is that the exact data $\x$ is eventually available to the decoder and is therefore only directly applicable to lossless compression. That is, bits-back is a solution to the lossless source coding problem but not a solution for reverse channel coding.

    \citeauthor{li2017strong} \cite{li2017strong} showed that is is possible to communicate $\z$ at an average coding cost of at most
    \begin{align*}
        I[\X, \Z] + \log_2(I[\X, \Z] + 1) + 5
    \end{align*}
    bits. That is, the coding cost is close to the information contained in $\z$. However, they also showed that in general it is not possible to significantly reduce this coding cost further. Even for optimal encoders and decoders we may therefore have to pay an overhead which is logarithmic in the mutual information. Note that this overhead is relatively small if the mutual information is large.

    One way to increase the mutual information (and thus reduce the relative overhead) is to communicate more information at once (for example, by bundling multiple frames of a video). Unfortunately, it can be computationally very expensive to do so. Agustsson \& Theis \cite{agustsson2020uq} showed that there is no general reverse channel coding algorithm whose computational cost is polynomial in the information content. If we want to transmit large amounts of information at once using as few bits as possible, then this may only be possible by spending a lot of computation. Nevertheless, some distributions can be communicated efficiently, both computationally and with low overhead.

    In the following, we will review two strategies for communicating stochastic information. One is a simple and efficient approach for simulating channels with additive uniform noise, and one is a general approach for communicating samples of arbitrary distributions. For a more thorough introduction to reverse channel coding, see Theis \& Yosri \cite{theis2021algorithms}.

    \subsection{Dithered quantization} \label{sec:uq}
    Consider a latent representation $\Z$\footnote{To reduce clutter, our notations here differs from those in Section~\ref{sec:neural-lossy-compression} on non-linear transform coding. Here, we denote the noise-injected latent representation by $\z$ (rather than $\tilde{\z}$), and denote transform coefficients by $\y$ (rather than $\z$). Our use of $\z$ here is consistent with Section~\ref{sec:rd-vae} and broader machine learning literature on latent variable models (see \cite{kingma2014vae, zhang2018advances}).} which is the output of a neural network followed by additive uniform noise, i.e.,
    
    \noindent
    \begin{align}
        \U &\sim \mathcal{U}([-0.5, 0.5)^n), \\
        \Y &= f(\X), \\
        \Z &= \Y + \U.
    \end{align}
    It turns out that we can efficiently communicate an instance of $\z$ using an old technique called \textit{dithered} or \textit{universal quantization} \cite{roberts1962noise}\cite{ziv1985universal}.
    
    Let $\U'$ be another vector of uniform noise independent of $\U$ which is available to both the encoder and the decoder. In practice this requirement can be achieved by generating noise using a pseudorandom number generator with the same random seed. For any fixed value of $\y = f(\x)$, it holds that \cite{ziv1985universal},
    \begin{align}
        \round{\y - \U'} + \U' \sim \y + \U. \label{eq:uq_dec}
    \end{align}
    That is, subtracting uniform noise, rounding, and then adding uniform noise back is distributionally equivalent to adding noise directly. We can exploit this for the communication of a uniform sample as follows. Define the random variable $\K = \round{\Y - \U'}$. Given a data sample $\x$, the encoder computes $\y = f(\x)$,  samples a value of $\bu'$, computes $\bk = \round{\y - \bu'}$, and entropy encodes $\bk$ into bits. The decoder
    receives $\bk$ and simply adds $\bu'$, and effectively obtains a sample of $\y + \U$ (by Eq.~\ref{eq:uq_dec}). To entropy encode $\bk$, we need to know its distribution. Assuming
    $\Z$ has marginal distribution $p_\Z$, we have \cite{zamir1992universal}
    \begin{align}
        \label{eq:uq_cond_entropy}
        P(\K = \bk \mid \U' = \bu') = p_{\Z}(\bk + \bu').
    \end{align}
    Note that we can condition on $\bu'$ to encode $\bk$ since it is known to the decoder. Eq.~\ref{eq:uq_cond_entropy} tells us that in order to encode $\bk$, we only need a model for the density of $\Z$. The expected coding cost is \cite{zamir1992universal,agustsson2020uq}
    \begin{align}
        \label{eq:uq_coding_cost}
        \mathbb{E}[-\log p_{\Z}(\K + \U')]
        &= \mathbb{E}[-\log p_{\Z}(\Y + \U)] \\
        &\geq h[\Z] \\
        &= h[\Z] - \cancelto{0}{h[\Z \mid \X]} \\
        &= I[\X, \Z]
    \end{align}
    with equality when $p_\Z$ is the true marginal distribution of $\Z$.
    
    This coding cost has two useful properties. First, it is equal to the amount of information transmitted (assuming $p_\Z$ models $\Z$ faithfully). That is, encoding $\K$ is a statistically efficient strategy for communicating a sample. Second, Eq.~\ref{eq:uq_coding_cost} is differentiable in $\y$ so that we can easily optimize an encoder using backpropagation. Agustsson \& Theis \cite{agustsson2020uq} exploited these facts to train neural encoders and decoders for images without the train-test mismatch commonly introduced by quantization.

    \subsection{Minimal random coding} \label{sec:importance-sampling}

    While dithered quantization can be computationally and statistically efficient, it is only able to communicate certain simple distributions. Several general algorithms have been developed to communicate a sample from arbitrary distributions \cite{harsha2007}\cite{cuff2008}\cite{cuff2013}\cite{li2017strong}\cite{havasi2018miracle}, though some have only been studied in the context of discrete distributions. Here we describe one algorithm based on importance sampling. In information theory, it is known as the \textit{likelihood encoder} \cite{cuff2013}\cite{song2016}. In machine learning, it has recently been introduced as \textit{minimal random coding} (MRC) \cite{havasi2018miracle}\cite{flamich2020cwq}.

    Assume both the encoder and decoder have access to $p(\z)$. For each given data observation $\x$, the encoder generates $N$ examples $\z_n \sim p(\z)$ and forms importance weights for a target distribution $q(\z \mid \x)$,
    \begin{align}
        w_n = \frac{q(\z_n \mid \x)}{p(\z_n)}.
    \end{align}
    It then randomly samples an index $k$ using the normalized importance weights, i.e.,
    \begin{align}
        P(n) = \frac{w_n}{\sum_{i=1}^N w_i}, \quad k \sim P.
    \end{align}
    The index is uniformly distributed and encoded using $\log_2 N$ bits. The decoder receives $k$ and reconstructs $\z_k$. It can do this if, for example, the encoder used a pseudorandom number generator to generate $\z_n$ with a random seed known to the decoder. The complexity of this procedure thus depends on the number of samples $N$ that needs to be generated.
    
    \citeauthor{havasi2018miracle} \cite{havasi2018miracle} showed that if the number of samples is
    \begin{align}
      N = 2^{\KL{q(\z \mid \x)}{p(\z)} + t}
    \end{align}
    and if the distribution of $\log q(\Z \,|\, \x)/p(\Z)$ is concentrated around its expected value, then $\Z_k$ quickly converges to $\Z$ (in a total variation sense) as $t$ increases. \citeauthor{havasi2018miracle} \cite{havasi2018miracle} applied MRC to neural network compression, while \citeauthor{flamich2020cwq} \cite{flamich2020cwq} used it for image compression.  Theis \& Yosri \cite{theis2021algorithms} showed that the coding cost of minimal random coding can be further reduced without any loss in quality.

    Cuff \cite{cuff2008} considered the setting where $m$ data points are encoded and communicated at once, and $p(\z)$ is the marginal distribution of $\Z$, that is, the average of the distributions $q(\z \mid \x)$ if we average over all $\x \sim p_{data}(\x)$.
    In this setting, he showed that if 
    \[
    N > 2^{m I[\X, \Z]},
    \]
    then the distribution of $(\Z_k, \X)$ converges to $q(\z \mid \x)p_{data}(\x)$ in total variation distance as $m$ goes to infinity. In other words, using on average $I[\X, \Z]$ bits, we can communicate a sample which approximately follows $q(\z \mid \x)$ in a total variation sense.

    \subsection{Stochastic versus deterministic coding}
    \label{sec:stochastic}

    Communication of information without any quantization requires a certain level of noise to be present. Without noise or quantization, there would be no limit to the amount of information we could send through the bottleneck of an autoencoder. This raises the question of whether a deterministic encoder with quantization or a stochastic encoder is better. \citeauthor{balle2021ntc} \cite{balle2021ntc} argue that we can always improve on dithered quantization with a deterministic encoder when performance is measured by a rate-distortion trade-off. Theis \& Agustsson \cite{theis2021stochastic} extended this argument to arbitrary stochastic encoders. That is, when we care about a rate-distortion trade-off, the best stochastic encoder is likely to perform worse than the best deterministic encoder. However, Theis \& Agustsson \cite{theis2021stochastic} also showed by example that when we additionally care about the realism of reconstructions (discussed in Section~\ref{sec:perceptual-losses}), stochastic encoders can perform significantly better than deterministic ones. Which one is better therefore depends on the setting of interest.
    
    \subsection{Lossy compression with diffusion}
    
    Reverse channel coding schemes enable novel compression schemes which deviate from the classical transform coding setup. One such approach is based on Gaussian diffusion models \cite{sohl2015deep}. These generative models learn the joint distribution of the data $\x = \z_0$ and copies of the data corrupted by increasing levels of noise,
    \begin{align}
        \z_{t + 1} = \alpha \z_t + \beta \mathbf{v}_t,
    \end{align}
    where $\mathbf{v}_t$ is Gaussian noise. For appropriate $\alpha$ and $\beta$, $\z_t$ as a function of $t$ can be viewed as approximating a Gaussian diffusion process.
    
    \citeauthor{ho2020denoising} \cite{ho2020denoising} first considered the rate-distortion performance of a scheme which would efficiently communicate an instance of $\z_t$ for some fixed $t$ before generating an estimate of $\x = \z_0$ with the help of the diffusion model. \citeauthor{theis2022diff} \cite{theis2022diff} demonstrated that this approach can work extremely well on small images when compared to the best transform coding schemes, especially when realism is considered. This is surprising considering that Gaussian noise is added directly to the data instead of applying an encoder transform first. Unlike neural transform coding which typically requires training many different models, the same diffusion model can be used to encode and decode a Gaussian sample at arbitrary bit-rates. The simplicity of this approach makes it very attractive from a theoretical perspective but its practicality is still an open question as the approach is very computationally expensive.
    
    Diffusion models have also been used in a transform coding setting by conditioning the generative model on the quantized output of an encoder. \citeauthor{saharia2021palette} employed a diffusion model for artefact removal on JPEG images \cite{saharia2021palette}, while \citeauthor{yang2022diff} \cite{yang2022diff} conditioned a diffusion model on a discrete latent variable in an end-to-end trained  variational autoencoder framework  \cite{yang2022diff}. These approaches do not require the as of yet computationally expensive communication of a sample but require dealing with a non-differentiable quantization operation as in other transform coding schemes. 
    They also do not benefit from the potential bit-rate savings of coding with a stochastic encoder \cite{theis2021stochastic}\cite{theis2022diff} (Section~\ref{sec:stochastic}).

    \section{Perceptual losses}
    \label{sec:perceptual-losses}
    Neural networks are only as good as the losses they are trained for. While in lossless compression the objective is clear -- namely to minimize the required number of bits to represent the data -- the story is a lot more complicated when we turn to lossy compression. Here we need to make decisions about which information to sacrifice in order to save additional bits. 
    In typical media compression applications, the goal is to make any reconstruction errors as imperceptible as possible, which raises complicated questions about how our brains perceive differences between signals.

    \subsection{Background}

    We can distinguish between two types of distortions, namely \textit{full-reference metrics} and \textit{no-reference metrics}. The former is a function which takes an image and its reconstruction as input while the latter only looks at the reconstruction to make a judgement about its quality. These can be motivated by two corresponding types of quality measures involving humans. \textit{Mean-opinion scores} (MOS) \cite{itu2016mos} are measured by having raters judge reconstructions on a scale from 1 (bad) to 5 (excellent). Since raters are only provided with the reconstructions, their judgments can be viewed as the output of a no-reference metric.  Many other \textit{image quality assessments} (IQAs) evaluate perceptual quality without reference to the original data \cite{storrs2018apc}, but MOS is the most widely used measure due to its simplicity. In contrast, \textit{degradation MOS} (DMOS) asks raters for a judgment based on both the unprocessed and the reconstructed data \cite{itu2016mos}, and thus is more akin to a full-reference metric.

    In addition to distortions, we may consider \textit{divergences} which depend on the probability distribution of the data and the marginal distribution of reconstructions. Driven mostly by the success of \textit{generative adversarial networks} (GANs) \cite{goodfellow2014gan} in producing realistic looking images, divergence optimization has become an important topic in neural compression.

    \subsection{Perceptual distortions}

    In this section we explore some of the metrics that are most likely to be encountered in the current literature on neural compression.
    However, there is a much larger body of potentially relevant work on IQAs and more broadly on perception that we will have to ignore here. For example, \textit{VMAF} \cite{liu2013fvqa}\cite{netflix2016vmaf} is a full-reference metric which is frequently used to evaluate the quality of video in industry but it has not yet found widespread use in the neural compression community. 
    
    Mean-squared error (MSE) and peak signal-to-noise ratio (PSNR) are frequently used in neural compression but do not predict perceived quality well \cite{eskicioglu1995iqa}. Another criterion widely used to evaluate images in machine learning and beyond is the structural similarity index (SSIM) \cite{wang2004ssim}. SSIM has been shown to correlate better with human visual perception and has been extensively studied and extended. The extension most commonly found in neural compression papers is the multi-scale SSIM (MSSSIM) \cite{wang2003msssim} which evaluates SSIM at multiple resolutions and combines them multiplicatively. The neurogram similarity index measure (NSIM) is another closely related metric which is applied to neurograms or spectrograms of audio signals \cite{hines2012nsim} and has been shown to correlate well with the perceived quality of speech \cite{hines2015visqol}.
    
    While MSE and PSNR are computed pixel-wise, SSIM is computed from small image patches. Let $\x$ and $\y$ be two aligned grayscale image patches extracted from an image and its reconstruction, respectively. Further, let $\mu_\x, \sigma_\x,$ and $\sigma_{\x\y}$ represent the average pixel value, the standard deviation and the covariance of pixel values as measured from these patches. Using these quantities, we define
    \begin{align}
        l(\x, \y) &= \frac{2 \mu_\x \mu_\y + C_1}{\mu_\x^2 + \mu_\y^2 + C_1}, \\
        c(\x, \y) &= \frac{2 \sigma_\x \sigma_\y + C_2}{\sigma_\x^2 + \sigma_\y^2 + C_2}, \\
        s(\x, \y) &= \frac{\sigma_{\x\y} + C_3}{\sigma_\x \sigma_\y + C_3},
    \end{align}
    where $C_1, C_2, C_3$ are positive constants to ensure numerical stability.
    The structural similarity function $s$ measures correlation, 
    while the luminance function $l$ and contrast function $c$ are chosen to respond to \textit{relative} changes in luminance or contrast. 
    These functions will not change much if, for example, both $\mu_\x$ and $\mu_\y$ are scaled by the same factor. This is consistent with Weber's law of how the human visual system perceives changes in these parameters \cite{wang2003msssim}.
    SSIM is defined as
    \begin{align}
        \text{SSIM}(\x, \y) = l(\x, \y)^\alpha c(\x, \y)^\beta s(\x, \y)^\gamma,
    \end{align}
    where $\alpha, \beta, \gamma$ are additional parameters which control the relative importance of the factors (typically set to 1). Note that $\text{SSIM}(\x, \x) = 1$.
    To compute a single value for an entire image, one approach is to use a sliding window (e.g., $8 \times 8$ pixels) and then to average SSIM values. MS-SSIM instead uses a smooth windowing approach to compute local statistics in order to avoid blocking artifacts \cite{wang2003msssim}. SSIM is defined for grayscale images. In the neural compression literature, SSIM is typically applied separately to RGB channels and the resulting values are averaged to evaluate color images.

    SSIM has been shown to perform significantly better at predicting human judgments than MSE on common distortions such as blurriness, noise or blocking artifacts \cite{wang2004ssim}. However, its limitations are also well documented and it tends to fail for reconstructions produced by generative compression approaches \cite{ledig2017srgan}\cite{mentzer2020hific}.

    ``When a measure becomes a target, it ceases to be a good measure'' \cite{strathern1997goodhart}. In line with this adage, directly optimizing neural networks for MS-SSIM offers mixed results when compared to MSE in terms of perceptual quality \cite{balle2018hyper}. Unlike many applications of IQA, a metric has to make meaningful predictions for all conceivable distortions to be useful as a target in neural compression and cannot have any blind spots. \citeauthor{ding2021iqa} \cite{ding2021iqa} recently compared a large number of IQA methods and found that many were unsuitable for direct optimization.
    
    A common approach to the design of more sophisticated distortions is to rely on deep neural networks which perform well in some other vision task. Typically, these distortions take the form
    \begin{align}
        \rho_\Phi(\x, \y) = \rho(\Phi(\x), \Phi(\y)),
    \end{align}
    where $\Phi$ is some representation derived from the hidden activations of a neural network and $\rho$ is typically the MSE. \citeauthor{gatys2016style} \cite{gatys2016style} compellingly demonstrated the ability of such metrics to capture semantic content at different levels of abstraction in their seminal paper on neural style transfer. \citeauthor{bruna2016superres} \cite{bruna2016superres} used distortions derived from VGG \cite{simonyan2015vgg} and scattering networks \cite{bruna2013scatter} and applied them to the task of super-resolution, which can be viewed as a simpler form of neural compression with a fixed encoder. They found that these distortions lead to sharper reconstructions than MSE but can also cause artifacts.

    \citeauthor{zhang2018lpips} \cite{zhang2018lpips} further investigated the efficacy of distortions based on VGG and found that they can significantly outperform SSIM on a range of artifacts including those generated by neural networks. They further proposed the \textit{learned perceptual image patch similarity} (LPIPS). LPIPS uses pretrained classifiers such as AlexNet \cite{krizhevsky2012alexnet} or VGG \cite{simonyan2015vgg} but its parameters are finetuned in a supervised manner to match human responses. As an alternative, \citeauthor{bhardwaj2020pim} \cite{bhardwaj2020pim} recently showed that representations learned in a completely unsupervised manner can be as effective as LPIPS and proposed the \textit{perceptual information metric} (PIM). Here, $\Phi$ was learned using a contrastive loss.
    
    As of this writing, no metric has reached the level of humans predicting the responses of other humans and how to close the gap is an open research question.
    Amir \& Weiss \cite{amir2021mmd} found that randomly initialized networks and a simple kernel-based metric can perform as well as VGG-based distortions in predicting human responses,
    raising interesting questions about the necessity and usefulness of neural representations in perceptual distortions.

    \subsection{No-reference metrics}
    \label{sec:noreference}

    While no-reference metrics are rarely used as training targets in the current neural compression literature, they are sometimes used for evaluation \cite{mentzer2020hific}. The \textit{natural image quality evaluator} (NIQE) \cite{mittal2013niqe}, for example, extracts nonlinear features from image patches sampled from a test image. It then fits a Gaussian distribution to these features and compares it to a Gaussian distribution fitted to features extracted from natural images. Many extensions of NIQE have been proposed and used in the IQA literature \cite{zhang2015ilniqe}.

    \textit{Deep IQA} \cite{bosse2016deepiqa} samples 32x32 image patches from a test image and uses a convolutional neural network to predict perceptual quality judgments. The predictions for different patches are averaged to yield a single score for an image. The parameters of the network were trained in a supervised manner on the LIVE dataset \cite{sheikh2006live} which only contains 981 distorted versions of 29 reference images.
    To augment this dataset and to reduce overfitting, \citeauthor{kim2019diqa} \cite{kim2019diqa} pretrained a neural network to predict pixel-wise distortions before training on the LIVE dataset.

    \subsection{Divergences and adversarial networks}
    An image compressor which always outputs a fixed image of high perceptual quality would score high in terms of any no-reference metric and consume zero bits. However, for a useful compressor of natural images, we expect a diverse set of reconstructions roughly following the distribution of uncompressed images. Such properties can be assessed with \textit{divergences}, measuring the deviation between the data distribution and the distribution of reconstructions. Requirements on a divergence $d$ are $d[p, q] \geq 0$ and $d[p, q] = 0 \iff p=q$ for all distributions $p$, $q$. A divergence does not need not be symmetric. Examples include the Kullback-Leibler divergence (KLD), the Jensen-Shannon divergence (JSD), or the total variation distance (TVD).
    
    Divergences are sometimes described as either \textit{zero-avoiding} or \textit{zero-forcing} \cite{minka2005divergences}. Zero-avoiding divergences assign a high penalty to models which assign zero probability to events ($q(\x) = 0$) which have positive probability under the reference distribution ($p(\x) > 0$). They are also called \textit{mode-covering} divergences since they encourage models to assign probability mass to all modes of a distribution.  
    Examples include the KLD or the $\chi^2$-divergence. On the other hand, zero-forcing divergences assign a high penalty to model distributions which assign positive probability ($q(\x) > 0$) to events which have zero probability ($p(\x) = 0$). These are also called \textit{mode-seeking} divergences since the resulting models tend to ignore some of the modes of distributions with multiple peaks. An examples is the reverse KLD (Eq.~\ref{eq:rkld} with $P$ and $Q$ switched). Zero-forcing divergences are especially useful for capturing realism since they discourage models from generating implausible reconstructions.

    We can give further motivation for the total variation distance (TVD). Instead of asking human observers to rate reconstructions as in a MOS test, consider an experiment where we randomly present either real data or reconstructions with equal probability, and ask the observers to make a binary decision as to whether the data shown is real \cite{denton2015gan}. An optimal observer will correctly classify the data with probability \cite{blau2018tradeoff}\cite{nguyen2009fdiv}
    \begin{align}
        p_\text{success} = \frac{1}{2} d_\text{TV}[q, p] + \frac{1}{2},
    \end{align}
    where $d_\text{TV}$ is the TVD,
    \begin{align}
        d_\text{TV}[q, p] = \int |q(\x) - p(\x)| \, d\x.
    \end{align}
    That is, minimizing TVD minimizes the chance of an optimal classifier correctly discriminating reconstructions from real data. Other divergences can be motivated by considering other losses and the associated risk of an optimal classifier \cite{nguyen2009fdiv}\cite{sriperumbudur2009ipm}.
    Optimizing divergences of high dimensional distributions is challenging. Nevertheless, approximations optimized by generative adversarial networks (GANs) have proven useful in practice \cite{goodfellow2014gan}\cite{nowozin2016fgan}.
    
    Weighted combinations of distortions and adversarial losses have produced very promising results in neural compression and related image reconstruction tasks \cite{ledig2017srgan}\cite{rippel2017waveone}\cite{santurkar2018generative}\cite{agustsson2019extreme}\cite{mentzer2020hific},
    \begin{align}
        \text{sup}_{D \in \mathcal{D}} \, \mathcal{L}_D(f, g) + \beta
        \mathcal{L}_{\rho}(f, g),
    \end{align}
    where $D$ is a discriminator network (adversary), and $(f, g)$ are a pair of encoder and decoder networks introduced in Section \ref{sec:3.2.1}. To give a simple example, we might choose
    \begin{align}
        \mathcal{L}_D(f, g) &:= \mathbb{E}[\log D(\X) + \log (1 - D(g(\Q{f(\X)}, \boldsymbol{\epsilon})))], \\
        \mathcal{L}_{\rho}(f, g) &:= \mathbb{E}[||\Phi(\X) - \Phi(g(\Q{f(\X)}, \boldsymbol{\epsilon}))||^2].
    \end{align}
    The feature representation $\Phi$ is typically a combination of pixels and VGG feature activations. 
    $\Q{\cdot}$ is a quantizer mapping the output of the encoder to a discrete number of values.
    In addition to the encoder's outputs, the decoder receives independent noise $\boldsymbol{\epsilon}$ as input.
    It can be shown that $\mathcal{L}_D(f, g)$ lower-bounds the Jensen-Shannon divergence between the data distribution and the marginal distribution of reconstructions \cite{goodfellow2014gan}. Similar losses can be used to target other divergences \cite{nowozin2016fgan}.
    For example, \citeauthor{agustsson2019extreme} \cite{agustsson2019extreme} used an LSGAN loss which targets a $\chi^2$-divergence \cite{mao2017lsgan} and were able to train autoencoders achieving much more detailed reconstructions at extremely low bit-rates than advanced classical codecs.

    Divergences are not just used for training but also for evaluating neural compression results. In particular, the Frechet inception distance (FID) \cite{heusel2017fid} measures the squared Wasserstein-2 distance between the data and reconstructions in the feature space of the \textit{inception-v3} network \cite{szegedy2015v3}, approximating both distributions as multivariate Gaussian.  Both FID and NIQE (Section~\ref{sec:noreference}) measure the distance between two Gaussian distributions. However, while NIQE estimates the distribution of features \textit{within} a single image, FID measures the distribution of features \textit{across} many different images.

    \subsection{Perception-distortion trade-off}

    An important question for neural compression is whether divergences are needed at all. Could we achieve the same results with just a distortion? Optimizing MSE, SSIM and even neural-network-based losses tend to produce artefacts but perhaps these will disappear as perceptual distortions improve. A strong counterargument to this
    hope was given by Blau \& Michaeli  \cite{blau2018tradeoff}. They showed that \textit{any} distortion is going to produce artefacts for some data unless the compressed representation allows us to reproduce the inputs perfectly. That is, for any lossy codec, the optimal reconstruction $\hatX$ of $\X$ optimized for the expectation of a given distortion will not preserve the data distribution  \cite[Theorem 1]{blau2018tradeoff}. 
    Note that this limitation does not apply if we are allowed to optimize divergences since then we can achieve perfect realism by requiring that all admissible decoders produce reconstructions such that $d[p_\hatX, p_\X] = 0$ for some divergence $d$, at least in theory.
    Blau \& Michaeli \cite{blau2018tradeoff} also showed that the achievable divergence can only increase when we demand lower levels of distortion. This makes sense, as the set of available decoders shrinks as we impose stronger constraints on them. Counterintuitively, however, this means that minimizing distortion can have the effect of reducing the realism of reconstructions.
    Blau \& Michaeli \cite{blau2018tradeoff} called this the \textit{perception-distortion trade-off}, that is, a small distortion limits the achievable realism and a low divergence (high realism) limits the achievable distortion. 
    However, the amount of tension between distortions and divergences depends on their particular choices \cite{dosovitskiy2016invert} \cite{ledig2017srgan} \cite{blau2018tradeoff}.

    \section{Task-oriented compression}\label{sec:task-oriented-compression}

    An increasing amount of data, such as climate data or satellite images, require efficient storage and transmission, and will only ever be ``seen'' by algorithms and machines that process them \cite{dubois2021lossy}. For such data, the ``perceptual'' quality of the reconstructions is irrelevant; rather, we only wish to preserve our performance on certain downstream tasks, when we use a compressed representation instead of the original data.  By performing downstream tasks directly on the compressed representation, and instead of a reconstruction in the data space, we can potentially achieve savings both in the bit-rate \cite{singh2020end}\cite{dubois2021lossy} as well as computational requirements of the system \cite{torfason2018towards}\cite{matsubara2021supervised}. 
    
    We may formalize \textit{task-oriented compression} as follows. Suppose we are provided $(\X, \Y)$ pairs from some joint distribution, where $\X$ needs to be compressed and sent to the receiver as before, while $\Y$ is available to both the sender and receiver, and has the interpretation of a \textit{target} or supervision signal for some task. A concrete example could be semantic segmentation, where $\x$ is an image, and $\y$ contains ground-truth semantic category labels for each pixel of $\x$. Given a sampled pair of data and target $(\x, \y)$, a task-oriented compression algorithm maps $\x$ to a representation $\z$, either deterministically or stochastically (in either case, $\z$ is another random variable constructed from $\x$). The transmission of $\z$ incurs some bit-rate loss $\R$, as usual. The decoder, having received $\z$, performs some downstream computation and evaluates performance given the target $\y$, resulting in a task-oriented distortion loss $\D$. Task-oriented compression then aims to to minimize a combination of the rate and task-oriented distortion losses, with different methods making different choices of $\R$ and $\D$.
    
    Arguably the earliest example is the Information Bottleneck principle \cite{tishby2000information}. Here the rate loss is given by the mutual information $\R = I[\X, \Z]$ as in the rate-distortion function (Section \ref{sec:lossy-background}), while the distortion is chosen to be a negative mutual information $\D = -I[\Z, \Y]$. The general idea is to encourage $\Z$ to become a minimal sufficient statistic of $\X$ for predicting $\Y$ \cite{alemi2016deep}. 

    Practical examples of task-oriented compression usually require $\Z$ to be discrete, so it can be compressed with the neural lossy compression approaches described in Section \ref{sec:neural-lossy-compression}. The rate loss is thus measured by the Shannon entropy, $\R = H[\Z]$, and the distortion is based on a concrete task, instead of an information-theoretic one as in the Information Bottleneck. Often, the decoder further computes a prediction $\haty$ given $\z$, e.g., by passing $\z$ through a neural network. The task distortion may be a conditional cross-entropy, $\D = H[\Y, \hatY | \Z]$, for a classification task\footnote{
    Let $p_{\Z \Y}$ denote the joint distribution of $\Z$ and $\Y$, and suppose the decoder makes a probabilistic classification $\haty \sim q_{\hatY|\Z=\z}$ for any given $\z$. The task loss for maximizing the likelihood of $\Y$ given $\Z$ is then $\D = \EX_{(\z, \y) \sim p_{\Z \Y}} [- \log q_{\hatY|\Z}(\y |\z)] = \EX_{\z \sim p_{\Z}}[H[p_{\Y|\Z=\z}, q_{\hatY|\Z=\z}]] =: H[\Y, \hatY | \Z]$, where $H[\cdot, \cdot]$ denotes (unconditional) cross entropy (defined in Eq.~\ref{eq:cross-entropy}).
    }, or a squared error loss, $\D= \EX[\| \Y - \hatY\|^2] = \EX[\| \Y - g(\Z)\|^2]$, for a regression task.
\citeauthor{singh2020end} \cite{singh2020end} considered the classification setting, where $\Z$ is a (often high-dimensional) feature tensor produced by the penultimate layer of a deep CNN. %
    They optimized for the bit-rate and classification accuracy by training end-to-end on the Lagrangian objective $\R + \lambda \D$, using the same additive uniform noise approximation as in \cite{balle2017end} (discussed in Section \ref{sec:ntc-with-uniform-quantization}). \citeauthor{matsubara2021supervised} \cite{matsubara2021supervised} adopted a similar end-to-end approach, but $\Z$ is chosen to be computed by an earlier layer of a neural network, due to a limited computational budget of the sender (e.g., an edge device sending a photo to a cloud server).
    \citeauthor{dubois2021lossy} \cite{dubois2021lossy} extended the task-oriented compression setup to consider multiple downstream tasks, aiming to learn a compressed representation that ensures good performance on a variety of tasks. A challenge is that we rarely know in advance all the downstream tasks of future interest, at compression time. They address this by focusing on tasks that are invariant under user-defined transformations to the input data $\x$, such as image classification under random translation or cropping.

    \section{Video compression}
    \label{sec:video-compression}
    
    While deep generative modeling has impacted image compression early on \cite{gregor2016towards}, its application to \textit{video} compression began more recently (circa 2018) due to the additional complexity in modeling videos as well as the increased computational complexity.

    A typical video codec consists of two steps: motion compensation and residual compression. The idea behind motion compensation is to predict the next frame in a video based on previous frames. 
    Traditionally, motion compensation relied on block motion estimation (i.e., matching entire patches in videos and memorizing compressed displacement vectors). By contrast, neural approaches typically 
    directly predict the displacement fields of pixels between adjacent frames. If the predicted displacement field is ``simple'', e.g., sparse, it can be efficiently compressed. Since certain aspects in a video can not easily be predicted, the (typically sparse) residual is separately compressed using image compression models (classical or neural).

    Recently proposed neural video codecs differ in design choices of the predictive model (e.g., stochastic vs deterministic) and the residual compression scheme (e.g., compression in latent space vs. pixel space). Another fundamental design choice is to either consider the low-latency setup in which a video has to be encoded and decoded on a frame-by-frame level, or the offline setup, in which case the video is encoded as a whole (using knowledge of future frames). For example, the offline setup may be adequate for video streaming services, while the online setup may be more suitable for video conferencing. Another line of research investigates hybridizing classical and neural codecs~\cite{ma2019image}. 

    One of the earliest approaches to joint prediction and residual compression based on convolutional architectures was  \cite{chen2017deepcoder}, which still used a traditional block-based motion estimation approach. For offline compression,  \cite{wu2018video}\cite{djelouah2019neural} formulated the video compression problem as a frame interpolation problem. Here, a subset of ``key frames'' are compressed as images, and the intermediate frames reconstructed based on video interpolation techniques.

\looseness=+1
    Most neural compression approaches currently focus on the low-latency (online) setup \cite{han2019deep}\cite{lu2019dvc}\cite{agustsson2020scale}\cite{golinski2020feedback}\cite{yang2020learning}\cite{lu2020end}. These approaches can be interpreted as autoregressive generative models for frame sequences \cite{yang2020hierarchical}. \citeauthor{han2019deep} \cite{han2019deep} and \citeauthor{habibian2019video} \cite{habibian2019video} first adapted the neural image compression framework of \citeauthor{balle2017end} \cite{balle2017end} to video data, framing it in a latent variable modeling context. While \citeauthor{han2019deep} \cite{han2019deep} proposed to encode the frame sequence using the predictive next-frame distribution of a stochastic recurrent neural network (eliminating the need to separately compress residuals) and adding a global latent variable that played a similar role as a key frame in traditional compression, \citeauthor{habibian2019video} \cite{habibian2019video} used 3D convolutions and explored structured priors such as PixelCNNs \cite{vandenoord2016pixelcnn} for lower bit-rates at the expense of increased runtime.

    In subsequent work, the separation of motion estimation and residual compression dominated and led to improvements over classical video codecs such as HEVC. \citeauthor{lu2019dvc} \cite{lu2019dvc} adopted a hybrid architecture that combined a pre-trained Flownet \cite{dosovitskiy2015flownet} and residual compression. \citeauthor{rippel2019learned} \cite{rippel2019learned} proposed a motion estimation approach with long-term memory and adaptive rate control.

    Another noteworthy innovation is scale-space flows. \citeauthor{agustsson2020scale} \cite{agustsson2020scale} used learned optical flows and warping to predict the next frame in sequence, however, it does so by adding a ``scale'' dimension to the optical flow field. This dimension allows the model to adaptively blur the source based on how well the next frame is predicted. 
    A general framework for low-latency video compression was recently introduced by \citeauthor{yang2020hierarchical} \cite{yang2020hierarchical}. The paper also showed that the frame reconstruction for models such as those of \citeauthor{agustsson2020scale} \cite{agustsson2020scale} could be improved by introducing a scale parameter that mediates between the autoregressive prediction and the compressed residual (in a similar way as how RealNVP improves over NICE \cite{dinh2014nice}\cite{dinh2016density}), as well as by using non-factorized priors.
    
    The components of neural video compression models are often strongly inspired by classical codecs, in particular those dealing with motion compensation. While this allows neural models to approximate the performance of their classical counterparts, it also introduces complexity and can limit their flexibility. Recent work was able to achieve state-of-the-art results using a greatly simplified approach dubbed the video compression transformer (VCT) \cite{mentzer2022vct}. Here, individual frames are independently transformed by encoder and decoder transforms but the latent representations are jointly encoded with a powerful entropy model based on transformers \cite{vaswani2017attention}.

    \chapter{Discussion and Open Problems}
    \label{sec:conclusion}
    
    Neural compression is a rapidly growing field and has made significant strides in both lossless and lossy compression. While neural approaches to image compression barely beat JPEG 2000 in 2016 \cite{theis2017cae}\cite{balle2017end}, they already outperformed the best known handcrafted codecs in 2018 \cite{zhou2018tucodec}.
    And in the \textit{Challenge on Learned Image Compression}\footnote{\url{http://compression.cc/}} of 2021, classical codecs did not even make it into the top 10. Similarly, the leading codec in the \textit{Large Text Compression Benchmark} relied solely on neural networks to predict text \cite{bellard2019nncp}.

\enlargethispage{-\baselineskip}    
    Nevertheless, many practical and theoretical and challenges remain.
    Chief among them is computational complexity, which stands in the way of the wider adoption of neural compression. 
    Computational feasibility is also a critical factor in the application of neural networks to new data modalities, such as point clouds and VR content. 
    There are also many open problems in neural lossy compression in particular, such as the design of loss functions and evaluation criteria, more efficient compression without quantization, and ways to mitigate the risk of miscommunication.

    A major roadblock for both neural lossless and lossy compression is high computational complexity.
    While neural-network-based approaches offer remarkable compression performance, they demand significantly higher computation than traditional codecs, and have so far mostly been developed in high-performance computing environments with GPUs. However, real-world applications, such as video streaming on mobile devices, come with stringent requirements on latency, power consumption, hardware compatibility, etc., at a much lower computation budget.  Neural compression methods will need to meet these requirements, and still deliver significant improvements over traditional methods, for them to be widely adopted \cite{minnen_current_2021}\cite{mukherjee_challenges_2022}. 
    
    New challenges arise from applying neural compression, or deep learning in general, to new types of data. Often, finding an effective digital representation of the data that meets application requirements is an art in itself, such as complex 3D scenes that need to be rendered from arbitrary viewing angles \cite{mildenhall2021nerf}\cite{ yu2021plenoctrees}\cite{muller2022instant}. The next, and often highly related question, is how to best process and ultimately compress such data with neural networks, e.g., in the non-linear transform coding paradigm. For example, point clouds can be compressed using voxelization followed by non-linear transform coding \cite{quach2019learning}\cite{guarda2019point}, but a naive implementation using a 3D convolutional autoencoder can quickly run out of memory, and care is needed to apply neural networks strategically \cite{guarda2019point}\cite{quach2020improved}. 
    In the case of image data, these basic issues around representation and neural architectures have been well addressed prior to the current wave of research in neural image compression: modern computers have long been able to efficiently represent and manipulate digital images as matrices, and the machine learning community have converged to convolution neural networks as the default architecture for image processing. For many emerging data types, these issues have yet to be fully resolved, and the solutions may involve substantial domain knowledge and well-designed data structures \cite{yu2021plenoctrees}\cite{muller2022instant}. We refer interested readers to the survey by \citeauthor{quach2022survey} \cite{quach2022survey} for a more in-depth discussion of these challenges in the context of point cloud compression.

    Many open questions also remain in the design of objective functions in lossy compression. For example, it is poorly understood to which extent divergences and adversarial approaches are needed for realism, or whether realism could also be achieved through well-crafted distortions and no-reference metrics. Adversarial losses also continue to pose challenges for tuning and optimization and no loss has yet emerged which can be trusted to reliably judge the perceptual quality of reconstructions in training and evaluation.
    On a related note, it remains to be seen how well neural networks can optimize various compression objectives compared to theoretical performance limits \cite{yang2022towards}\cite{wagner2021neural}, and to better understand the cause of suboptimality.

    The use of quantization continues to cause a mismatch between training and test time performance, and how much this affects compression performance is still not clearly understood. Reverse channel coding is a promising alternative which eliminates the need for quantization, 
    but has only recently been considered in neural compression \cite{theis2021algorithms}.
    Open questions include the design of efficient coding schemes and the impact these schemes have on performance when compared to approaches based on quantization.

    With the advances in neural lossy compression also emerges the risk of miscommunication, especially in semantically constrained domains. The reconstructions from neural lossy compression models, especially those targeting realism \cite{rippel2017waveone}\cite{agustsson2019extreme}\cite{mentzer2020hific}\cite{theis2022diff}, can appear highly realistic at extremely low bit-rates, yet misrepresent the  semantic content or other ``relevant'' information in the original data. The reconstructions might also be stochastically generated, and differ across different users and times of access. 
    These effects raise safety and ethics concerns where such miscommunication can have severe consequences, e.g., in the transmission of security camera videos. 
    Besides building safeguards into our methods, addressing these concerns may also require re-examining our choice of objective functions, for example,
    the distortion function and its effectiveness at capturing what is truly ``relevant'' to the end user of lossy compression.

\begin{acknowledgements}
The authors thank Karen Ullrich, Yingzhen Li, Thanasi Bakis, and David Minnen for providing valuable feedback on our manuscript. Yibo Yang acknowledges support from the Hasso Plattner Research School at UC Irvine. Stephan Mandt acknowledges support by the National Science Foundation (NSF) under the NSF CAREER Award 2047418; NSF Grants 2003237 and 2007719, the Department of Energy, Office of Science under grant DE-SC0022331, as well as gifts from Intel, Disney, and Qualcomm. 
\end{acknowledgements}

\backmatter  %

\printbibliography

\end{document}